\documentclass[journal]{IEEEtran}

\ifCLASSINFOpdf
\else
\fi
\usepackage{amsmath}
\usepackage{amssymb}
\usepackage{bm}
\usepackage{algorithm}
\usepackage{algorithmicx}
\usepackage{algpseudocode}
\usepackage{array}
\usepackage{caption}
\usepackage{subcaption}
\usepackage{graphicx}
\usepackage{multirow}
\usepackage{color, colortbl}

\usepackage[utf8]{inputenc}% utf8, for example

\definecolor{Orange}{RGB}{255, 219, 153}

\newcolumntype{a}{>{\columncolor{white}}c}
\newcolumntype{b}{>{\columncolor{Orange}}c}

\algnewcommand\algorithmicinput{\textbf{INPUT:}}
\algnewcommand\INPUT{\item[\algorithmicinput]}

\begin{document}

\title{Distribution-based Label Space Transformation for Multi-label Learning}

\author{Zongting Lyu, Yan Yan, and Fei Wu}

\markboth{Journal of \LaTeX\ Class Files,~Vol.~14, No.~8, August~2015}%
{Shell \MakeLowercait{et al.}: Bare Demo of IEEEtran.cls for IEEE Journals}

\maketitle

\begin{abstract}
Multi-label learning problems have manifested themselves in various machine learning applications. The key to successful multi-label learning algorithms lies in the exploration of inter-label correlations, which usually incur great computational cost. Another notable factor in multi-label learning is that the label vectors are usually extremely sparse, especially when the candidate label vocabulary is very large and only a few instances are assigned to each category. Recently, a label space transformation (LST) framework has been proposed targeting these challenges. However, current methods based on LST usually suffer from information loss in the label space dimension reduction process and fail to address the sparsity problem effectively. In this paper, we propose a distribution-based label space transformation (DLST) model. By defining the distribution based on the similarity of label vectors, a more comprehensive label structure can be captured. Then, by minimizing KL-divergence of two distributions, the information of the original label space can be approximately preserved in the latent space. Consequently, multi-label classifier trained using the dense latent codes yields better performance. The leverage of distribution enables DLST to fill out additional information about the label correlations. This endows DLST the capability to handle label set sparsity and training data sparsity in multi-label learning problems. With the optimal latent code, a kernel logistic regression function is learned for the mapping from feature space to the latent space. Then ML-KNN is employed to recover the original label vector from the transformed latent code. Extensive experiments on several benchmark datasets demonstrate that DLST not only achieves high classification performance but also is computationally more efficient. 
\end{abstract}


\IEEEpeerreviewmaketitle

\section{Introduction}

\IEEEPARstart{M}{ulti-label} learning naturally arise in various machine learning tasks such as text mining \cite{Schapire:ML2000}, image classification~\cite{Boutell:PR2004} and annotation~\cite{Chen:FastTag2013, Wang:Sparse2009} and bioinformatic analysis~\cite{Barutcuoglu:Bio2006}. For example, a document may be annotated with multiple diverse tags, an image may contain multiple object categories, a gene is usually multifunctional. As a generalization of multi-class learning, multi-label learning allows each instance to be assigned to a set of labels rather than a single label. Given potential applications in a variety of real-world problems such as keyword suggestions~\cite{Agrawal:KeywordSuggestion2013} and video segmentation~\cite{Snoek:VideoSegment2006}, multi-label learning, especially multi-label classification has been extensively studied and is attracting more research attention.

\begin{figure}[t]
\centering
\includegraphics[width=\linewidth]{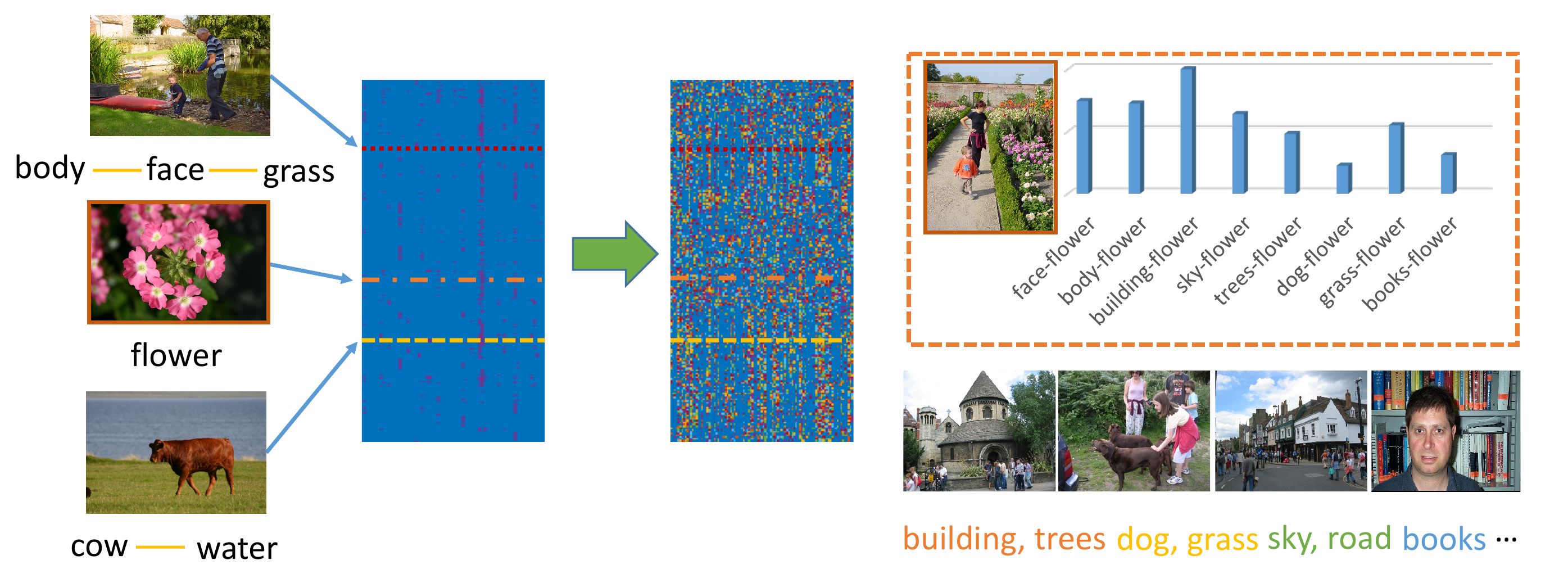} 
\caption{An illustration of label transformation of the proposed method via distribution alignment. Most methods disable when the input label vectors are very sparse, either resulting from large label vocabulary or missing labels. By employing label similarity based distribution, the proposed method exploits comprehensive label correlations. For example, \textit{flower} seldom co-occurs with \textit{building, dog, books, etc.,} in the training images, nonetheless its association with these labels can still be well captured by describing the distribution defined on pairwise label similarity. Then, the distribution information is further encoded in the latent dense label representation.}
\label{fig:Label transform}
\end{figure}

Well-established multi-label classification methods roughly follow two lines of research~\cite{Tai:PLST2012, Li:CVPR2016, Zhang:Review2014}, namely \textit{Algorithm Adaptation} and \textit{Problem Transformation}. Algorithm adaptation methods adopt certain existing algorithms and adapt them to solve the multi-label classification problem. Representative members of this group include Rank-SVM~\cite{Elisseeff:RankSVM2001}, ML-KNN~\cite{Zhang:MLKNN2007} and Instance based Logistic Regression~\cite{Cheng:IBLR2009}. 

Problem transformation methods, on the other hand, reformulate the multi-label classification problem into one or more sub learning tasks. Typical examples include Binary Relevance (BR), Classifier Chains (CC)~\cite{Read:CC2011, Liu:NIPS2015}, Label Ranking (LR)~\cite{Funkranz:LabelRank2008}, Label Powerset (LP)~\cite{Kong:LabelSet2013} and its variants such as Random k-Label Sets~\cite{Tsoumakas:RAKEL2007} and Pruned Problem Transformation~\cite{Read:PPT2008}. BR decomposes multi-label classification to many separate single-label binary classification tasks, each for one of the labels. CC takes the label dependency into consideration by constructing a chain of binary classifiers, where each classifier additionally leverage the previous labels as its input feature. LR reformulates the multi-label classification problem into a task of ranking labels on hand by relevance and determining the threshold of the relevance. LP reduces multi-label classification to multi-class classification by treating each observed label set as a distinct multi-class label. The problem transformation approaches are more advantageous since any algorithm to the transformed task can be used to solve multi-label classification problems.

More recently, research on multi-label classification generally fall into two learning paradigms~\cite{Tai:PLST2012, Li:CVPR2016}. The first is \textit{structured output learning} paradigm, which focuses on modeling label structure and exploiting inter-label correlations, then using them to predict label vector for test instances~\cite{Hariharan:ICML2010, Li:CVPR2016}. Label correlations are typically encoded in a graph structure, such as ChowLiu Tree~\cite{Chow:TIT1968} and Maximum Spanning Tree~\cite{Li:UAI2014}, and                             conditional label structure can be approximately learned via Structured Support Vector Machine and Markov Random Field. To further assist the label structure learning, ~\cite{Jiang:TPG2012, Kong:KDD2013, Li:CVPR2016} explicitly incorporate feature information into the learning process. However, such label structure learning methods are computationally expensive. 

The second model employs \textit{label space reduction}~\cite{Zhang:ICML2012}, which encodes the original label space to a low-dimensional latent space either through random projection~\cite{Zhou:ML2012}, canonical correlation analysis (CCA) based projection or by directly learning the projected codes~\cite{Lin:FAIE2014}. Subsequently, prediction is performed on the low-dimensional latent space, whose results are translated back to the original label space via a decoding process, thus the original labels for test instances can be recovered. Moreover, algorithms with both label space and feature space dimension reduction have been proposed, such as conditional principal label space transformation~\cite{Chen:CPLST2012}. In addition,~\cite{Jing:SLRM2015, Yu:LEML2014} take a more direct approach by formulating the label prediction problem as learning a low-rank linear mapping from feature space to label space. However, these methods usually suffer from information loss and depend on the reduced dimension of the latent space.

In recent years, the proliferation of labels pose great challenge to existing multi-label learning methods. Due to the large label vocabulary, the label vectors are usually characterized by high dimensionality and remarkable sparsity. The sparsity originates from two sources: (1) For each instance, only a small number of labels are present, namely, the label vector has little support (\textit{Sparsity I}). (2) For a certain set of labels, very few training instances are assigned to it (\textit{Sparsity II}). In the following paper, we refer to these two types of sparsity as label set sparsity and training data sparsity respectively. Although the label space dimension reduction approaches target to address this problem, the performance of the reduced latent space is not satisfying in terms of prediction accuracy and computational complexity. The reason is that the dimension reduction process in these methods may incur information loss such that the original inter-label correlations will not be fully preserved in the latent space.  

Following the general label space transformation framework proposed in~\cite{Tai:PLST2012}, we propose a novel distribution-based label space transformation model (DLST). By aligning the distribution between the original label space and the latent space, an optimal transformed code can be learned for each label vector. The advantages of employing distribution alignment in label space transformation lie in two aspects: (1) In contrast to conventional approaches which lose information, extra information that are beyond the original label vector can actually be fulfilled in the latent code according to the distribution. As a result, more complicated label correlations can be captured and approximated in the latent code. (2) In face of label set sparsity and training data sparsity, the proposed model is still able to recover the whole distribution.  It has been empirically verified that the number of labels per instance required for DLST to obtain highest score is far less than comparing baselines. Similar phenomenon is also observed for the number of training instances per class required to achieve the best performance.

 As shown in Figure~\ref{fig:Label transform}, the latent codes derived by the proposed method are much denser than original label vectors which also preserve the distribution of the original label space. Therefore, it can be expected that the classification performance using the transformed latent codes will be significantly better than the original sparse labels. 

In the training phase, a regression function is learned to map the original data in the feature space to the transformed code in the latent space. For each test instance, the corresponding latent code can be computed by applying the regression function. Then, ML-KNN~\cite{Zhang:MLKNN2007} is employed as the decoder to recover the original label vector from the latent code of each instance. The proposed model can also be extended with kernel tricks to deal with nonlinear regression from original feature to the latent code.

Since the proposed model is capable of tackling label set sparsity and training data sparsity. Real-world examples of these two types of sparsity are missing data and limited training data. The performance of DLST is relatively stable across varying missing ratios or training data ratios. This can be attributed to the distribution used in DLST. Rather than limited to the given label vectors, DLST captures the whole distribution of the label space by fitting the observed label vectors using a distribution with maximal variance, and transmits the distribution to the latent space. 
 
The contributions of this paper are:
\begin{itemize}
\item The proposed method takes advantage of distribution to capture more comprehensive inter-label correlations in the original label space and transmits it to the latent space. 
\item The dense latent code learned by DLST successfully addresses label set sparsity by distinguishing concurrent label patterns from most other unrelated labels.
\item The proposed model effectively alleviates the requirement on training data size to achieve high multi-label classification performance.  
\end{itemize}

\section{A Distribution based Multi-Label Learning Framework}

\begin{figure*}[t]
\centering
\includegraphics[width=\linewidth]{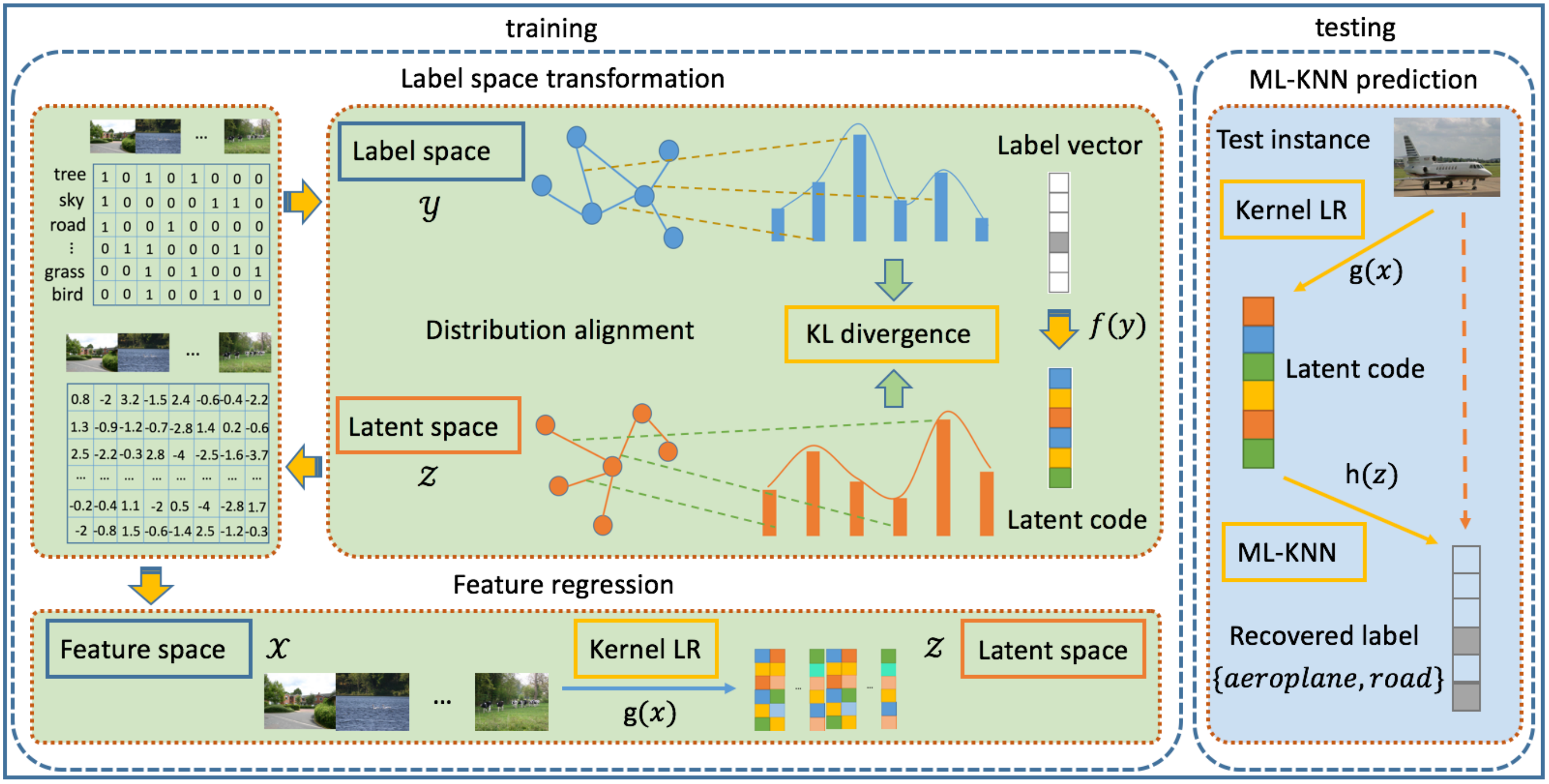}
\caption{Schematic illustration of DLST. In the training phase: Label space transformation and Feature regression. Label space transformation converts sparse label vector into dense latent code while preserving the distribution information of the original label space. Feature regression learns a kernel logistic regression function to map image features to latent codes. In the testing phase, the latent code of test instance is first learned by applying the regression function, and then ML-KNN prediction is performed to recover the original label vector.}
\label{fig:Schematic}
\end{figure*}

\subsection{Preliminaries}
Let $\mathcal{D}=\{(\bm{x}_i, \bm{y}_i)\}_{i=1}^n$ denote the set of labeled training data, where each instance $\bm{x}_i \in \mathcal{X} \subset \mathbb{R}^d$ is associated with a subset of $K$ possible labels represented by a binary vector $\bm{y}_i \in \mathcal{Y} \subset {\{0,1\}}^K$. $y_{ij}=1$ when $\bm{x}_i$ is assigned the $j$-th label and 0 otherwise. The goal of multi-label classification is to learn a mapping $\mathcal{M}: \mathcal{X} \rightarrow \mathcal{Y}$ for predicting the label vector of each test instance. For convenience of notation, the feature vectors and label vectors of training instances are arranged in row to form the input feature matrix $\bm{X} \in \mathbb{R}^{n \times d}$ and label matrix $\bm{Y} \in \mathbb{R}^{n \times K}$.  

Traditional multi-label classification methods aim to learn a binary classifier $\mathcal{M}_j: \mathcal{X} \rightarrow \mathcal{Y}_j$ for each dimension of the label vector, where $\mathcal{Y}_j$ indicates the $j$-th dimension of the original label space. In the case of numerous labels, it will become computationally prohibitive for these methods to predict the label vector. 

To tackle this challenge, a novel label space transformation learning framework was proposed, where each label vector $\bm{y}_i$ is firstly encoded into a point $\bm{z}_i \in \mathcal{Z} \subset \mathbb{R}^r$ in the latent space with an encoding process $f(\bm{y}): \mathcal{Y} \rightarrow \mathcal{Z}$. Similarly, the latent codes are also stacked in row to form a matrix $\bm{Z} \in \mathbb{R}^{n \times r}$. Then, the multi-label classification problem on $(\bm{X}, \bm{Y})$ turns to a multi-dimensional regression problem on $(\bm{X}, \bm{Z})$. After obtaining the regression function $g(\bm{x}): \mathcal{X} \rightarrow \mathcal{Z}$ that has a high prediction accuracy on $\bm{z}$, the framework will then map $g(\bm{x})$ back to the original label space via some decoder $h(\bm{z}): \mathcal{Z} \rightarrow \mathcal{Y}$. The whole learning process is illustrated in Figure~\ref{fig:Schematic}. Note that with the transformed latent codes $\bm{Z}$, the regression process $g(\bm{x})$ is open to be replaced by any mapping algorithm from data feature to multi-label vectors. 

\subsection{Label Space Transformation}
In this section, we propose an implicit encoding~\cite{Lin:FAIE2014} module by aligning the distribution of label space $\mathcal{Y}$ and that of latent space $\mathcal{Z}$. While traditional multi-label learning methods which explicitly model the inter-label correlations~\cite{Bi:AAAI2014, Bradley:ICML2010, Li:UAI2014, Tan:CVPR2015, Zhang:SIGKDD2010, Zhu:Spatial2017, Wang:CNN-RNN2016}, the depth of the correlations that they investigated are no more than three order limited by the computational complexity. Moreover, the correlations captured in the existing models are either local or global, and have a bias towards similarity between labels while dissimilarity is mostly ignored. 

In contrast, our model provides a more comprehensive description of the correlations between labels by employing distribution. Firstly, the similarities in the label space are transformed into probability distribution, and approximated by the distribution of latent code in a reduced space. Based on this distribution alignment, a new label representation can be learned which fills out the original label distribution information. As a result, the proposed model is not limited to the number of given label vectors and their completeness, which greatly expands its flexibility.

To derive the distribution $Q$ of label vectors $\bm{Y}$ for the labeled training instances, first define $q_{ij}$ as the probability of observing the similarity between label vectors $\bm{y}_i$ and $\bm{y}_j$ among all pairs of labeled training instances. Following t-SNE~\cite{Maaten:tSNE2008}, we utilize a Student t-distribution with one degree of freedom to transform the Euclidean distances into probabilities, as shown below 

\begin{equation}
\label{eq:label distribution}
Q_{ij}=\frac{{(1+d(\bm{y}_i,\bm{y}_j))}^{-1}}{\sum_{l \neq m}{(1+d(\bm{y}_l,\bm{y}_m))}^{-1}}
\end{equation}

Let $U$ denote the distribution of instances with to-be-learnt representations $\bm{Z} \in \mathbb{R}^{n \times r}$ in the latent space, then $U$ can be calculated as follows

\begin{equation}
\label{eq:latent distribution}
U_{ij}=\frac{{(1+d(\bm{z}_i,\bm{z}_j))}^{-1}}{\sum_{l \neq m}{(1+d(\bm{z}_l,\bm{z}_m))}^{-1}}
\end{equation}
 where $r$ is the dimension of the latent space. 
 
 To approximate the distribution $Q$ with that of latent code $\bm{Z}$ in the reduced space, we adopt the Kullback-Leibler divergence to measure the distribution discrepancy between $Q, U$, which can be formulated by   

\begin{equation}
\label{eq:objective 1}
\min_{\bm{Z}} \ D(Q_{ij}||U_{ij})=\min_{\bm{Z}} \sum_{i \neq j}q_{ij}log\frac{q_{ij}}{u_{ij}}
\end{equation}

This KL-divergence based distribution alignment technique considers the comprehensive correlations among the label vectors. The underlying assumption is that instances with highly correlated label vectors tend to have high similarity in the input data space. Therefore, instances with the same labels tend to be drawn much closer in the latent space.

\subsection{Feature Regression}

With the optimal latent representation $\bm{Z}^*$, the original multi-label classification problem on $(\bm{X}, \bm{Y})$ converts to a multi-dimensional regression problem on $(\bm{X}, \bm{Z}^*)$. The mapping function $g(\bm{x})$ is actually open for any effective multi-label prediction models, such as linear regression, ridge regression and logistic regression. More generally, any algorithm that learns a mapping from data features to the multi-label vectors can be exploited here, with a boosted performance than the original algorithm.

In this paper, we use kernel logistic regression to learn the mapping from features to latent codes. The reason are twofold: on one hand, logistic regression not only outputs the predictions but also the corresponding probabilities for the prediction. On the other hand, it can easily be extended to a kernelized version where nonlinear mappings are included. 

In kernel logistic regression, each instance $\bm{x}_i$ is mapped to the Reproducing Kernel Hilbert Space (RKHS) as $\bm{\phi}(\bm{x}_i) \in \mathbb{R}^{v}$, which also form a kernel feature matrix $\bm{\Phi} \in \mathbb{R}^{n \times v}$. In RKHS, the inner product between kernel features can be efficiently calculated by applying kernel trick $\kappa(\bm{x}_i, \bm{x}_j)$, where $\kappa$ is the introduced kernel function. Employing non-linear kernel functions, the linear mapping from kernel features to the latent codes are actually non-linear mappings from the original feature space to the latent space. In this paper, we treat each dimension of the latent code separately and learn a linear mapping $g_j(\bm{x})$ in RKHS for the $j$-th dimension. The objective function of kernel logistic regression is as follows

\begin{equation}
\label{eq:kernel LR}
\min \sum_{i=1}^n {log(1+e^{-z^*_{ij}\bm{\phi}(\bm{x}_i)g_j})}+\lambda ||g_j||_2^2
\end{equation}
where $z_{ij}$ is the $j$-th entry in $\bm{z}_i$, and $\lambda$ is a weighting parameter. 

Following the common practice in literature, let $g_j$ fall in the span of the kernel features for training instances, i.e. $g_j=\bm{\Phi}^{\top}\bm{c}_j$ with $\bm{c}_j$ as the spanning coefficients. Then in Eq.~\eqref{eq:kernel LR}, $\bm{\phi}(\bm{x}_i)g_j=(\bm{\phi}(\bm{x}_i)\bm{\Phi}^{\top})\bm{c}_j$. It can be seen that the training cost of kernel logistic regression is positively related to the training set size $n$, where is undesirable for large-scale datasets.

Note that not all training instances are required to form the span, as redundancy may exist between kernel features of training instances. Therefore, we only sample a small part of them for building the kernel feature matrix and use it as the basis to span the $j$-th mapping $g_j$. Hence, $g_j=\bar{\bm{\Phi}}^{\top}\bar{\bm{c}}_j$ where $\bar{\bm{c}}_j \in \mathbb{R}^s$ is the coefficients that need to be learned and $s$ denotes the sampling size. Then the training cost of kernel logistic regression can be greatly reduced, making it more efficient for training as well as predicting. The specific sampling strategy can be either random sampling or other more sophisticated methods.

\subsection{Multi-Label Prediction} 

For a test instance $\bm{x}_t$, based on the learned regression function $g_j$, the $j$-th dimension of the latent code $\bm{z}_t$ for $\bm{x}_t$ can be forecasted. In addition, the probabilities of $z_{tj}$ can be obtained as follows

\begin{equation}
\label{eq:test code}
p(z_{tj}|\bm{x}_t) = {(1+e^{-z^*_{ij}\bm{\phi}(\bm{x}_i)g_j})}^{-1}\end{equation}

To obtain the label vector for each test instance $\bm{x}_t$, the latent code learned in the previous step needs to be further mapped back to the original label space $\mathcal{Y}$ through some decoder $h(\bm{z})$. In this paper, ML-KNN is employed to recover the original label for test instances.

For each test instance $\bm{x}_t$, ML-KNN first identifies its $k$ nearest neighbors $N(t)$ in the training set. Then, based on the label sets of these neighbors, the label vector $\bm{y}_t$ for $\bm{x}_t$ can be determined using the following maximum a posteriori principle

\begin{equation}
\label{eq:test label}
y_{tj} = arg\max_{b \in \{0, 1\}} p(A_b^j|C_{m_j}^j)
\end{equation}
where $A_1^j$ indicates that instance $\bm{x}_t$ has label $j$, while $A_0^j$ denotes that $\bm{x}_t$ is not assigned label $j$. $C_m^j$ denotes that among the $k$ nearest neighbors of $\bm{x}_t$, there are exactly $m$ instances which are assigned the $j$-th label, $m$ can be calculated by $m_j = \sum_{a \in N(t)} y_{aj}$. Using Bayesian rule, Eq.~\eqref{eq:test label} is equivalent to the following objective function

\begin{equation}
\label{eq:objective 2}
y_{tj} = arg\max_{b \in \{0, 1\}} p(A_b^j)p(C_{m_j}^j|A_b^j)
\end{equation}

As shown in Eq.~\eqref{eq:objective 2}, in order to determine the label vector $\bm{y}_t$, all the information needed is the prior probabilities $p(A_b^j)$ and the posterior probabilities $p(C_{m_j}^j|A_b^j)$, which can all be directly estimated from the training instances. Problem~\eqref{eq:objective 2} can be similarly solved as in~\cite{Zhang:MLKNN2007}.

\begin{algorithm}[h]
\caption{The latent representation optimization for problem~\eqref{eq:objective 1}}
\label{alg:algorithm 1}

\begin{algorithmic}[1]
\INPUT Data matrix $\bm{X}$, label matrix $\bm{Y}$, $\epsilon = 10^{-6}$.
\Require optimization parameters: learning rate $\eta$, momentum $\alpha(t)$, number of iterations $T$.
\State $\bm{Z}=\bm{0}$
\State Compute the probability distribution $Q_{ij}$ of the original label space according to Eq.~\eqref{eq:label distribution}

\While{$||\bm{Z}^{t+1}-\bm{Z}^{t}||^2 \ge \epsilon$}
\State Update the distribution $U_{ij}^t$ according to Eq.~\eqref{eq:latent distribution}
\State Update gradient according to Eq.~\eqref{eq:optimization 1}
\State Update $\bm{Z}$ according to Eq.~\eqref{eq:update Z}
\EndWhile \\
\Return $\bm{Z}^*$

\end{algorithmic}
\end{algorithm}

\begin{algorithm}[h]
\caption{Multi-label propagation for problem~\eqref{eq:objective 2}}
\label{alg:algorithm 2}

\begin{algorithmic}[1]
\INPUT Original data matrix $\bm{X}$, Original label matrix $\bm{Y}$, $\epsilon = 10^{-6}$.

\State Training:
\State Reduce the original label matrix $\bm{Y}$ to a latent space with code matrix $\bm{Z}^*$ via algorithm~\ref{alg:algorithm 1}. 
\State Learn a kernel logistic regression $g(\bm{x})$ from $\bm{X}$ to $\bm{Z}^*$ by solving problem~\eqref{eq:kernel LR}.

\State Prediction: 
\State For each test instance, derive the latent code $g(\bm{x}_t)$.  
\State Map $g(\bm{x}_t)$ to label space, and recover the label vector of test instances according to Eq.~\eqref{eq:objective 2}.

\Return $\bm{Y}_t$

\end{algorithmic}
\end{algorithm}

\subsection{Optimization}

The objective function of problem~\eqref{eq:objective 1} is non-convex, thus only local optimum can be obtained. Since problem~\eqref{eq:objective 1} is an unconstrained optimization problem, to learn a locally optimal $\bm{Z}$, we propose to exploit gradient descent based optimization methods.  From Eq.~\eqref{eq:objective 1}, we have 

\begin{equation}
\label{eq:objective}
\min_{\bm{Z}} \ \sum_{i \neq j} Q_{ij}log{Q_{ij}}-Q_{ij}log{U_{ij}}
\end{equation}

Since $Q_{ij}$ solely depends on the labels of training data and remains fixed during the optimization procedure, therefore, problem~\eqref{eq:objective} can be reduced to 

\begin{equation}
\label{eq:reduced}
\min_{\bm{Z}} \ \mathcal{L}=\sum_{i \neq j}-Q_{ij}log{U_{ij}}
\end{equation}

Combining Eq.~\eqref{eq:label distribution} and ~\eqref{eq:latent distribution}, the gradient of Eq.~\eqref{eq:reduced} w.r.t. $\bm{Z}$ can be derived as follows

\begin{multline}
\label{eq:optimization 1}
\frac{\partial \mathcal{L}}{\partial \bm{Z}}
=\frac{-Q_{ij}}{U_{ij}} \cdot 
\frac{\partial U_{ij}}{\partial \bm{Z}} \\
 ={(1+||\bm{z}_i-\bm{z}_j||^2)}^{-1}(\bm{z}_i-\bm{z}_j) \\
-\frac{\sum_{l \neq m}{(1+||\bm{z}_l-\bm{z}_m||^2)}^{-2}(\bm{z}_l-\bm{z}_m)}{\sum_{l \neq m}{(1+||\bm{z}_l-\bm{z}_m||^2)}^{-1}}\end{multline}

With gradients calculated in Eq.~\eqref{eq:optimization 1}, effective gradient descent based optimization methods can be further applied to derive optimal $\bm{Z}$. The update strategy of $\bm{Z}$ is as follows

\begin{equation}
\label{eq:update Z}
\bm{Z}^{t+1}=\bm{Z}^{t}+\beta \frac{\partial \mathcal{L}_1}{\partial \bm{Z}}+\alpha(t+1)(\bm{Z}^{t}-\bm{Z}^{(t-1)})
\end{equation}
where $\bm{Z}^{t}$ denotes the optimal $\bm{Z}$ at $t$-th iteration, $\beta$ is the learning rate, $\alpha(t)$ is the momentum at $t$-th iteration. The stopping criteria for the algorithm is $||\bm{Z}^{t+1}-\bm{Z}^{t}||^2 \le \epsilon$, with a maximum iteration number $5000$. The details of the encoding algorithm are presented in Algorithm~\ref{alg:algorithm 1}. Algorithm~\ref{alg:algorithm 2} summarizes the whole procedure of DLST. The complexity of the proposed algorithm is $\mathcal{O}(n^2)+\mathcal{O}(s)+\mathcal{O}(n)$, where $n$ is the number of labeled training data, and $s$ is the sampling size for kernel logistic regression. 

\begin{table}[h]
\centering
\begin{tabular}{l|c|c|c|c|r}
\hline \hline
Dataset & type & n & d & K & card  \\ \hline
\rowcolor{Orange}
Scene & image & 2,407 & 294 & 6 & 1.074  \\
 \hline
Emotions & music & 593 & 72 & 6 & 1.869  \\
 \hline
\rowcolor{Orange}
Yeast & biology & 2,417 & 103 & 14 & 4.237  \\
 \hline
Mediamill & video & 43,907 & 120 & 101 & 4.376  \\
 \hline
\rowcolor{Orange}
MSRC & image & 591 & 512 & 23 & 2.508 \\
 \hline
SUNattribute & image & 14,240 & 512 & 102 & 15.526 \\
 \hline
\end{tabular}
\caption{Dataset statistics used in the experiments. n is the number of instances; d is the dimensionality of instances; K is the number of possible labels; card is the average number of labels per instance.}
\label{tab:dataset statistics}
\end{table}

\begin{figure*}[t]
\centering
\includegraphics[width=\linewidth]{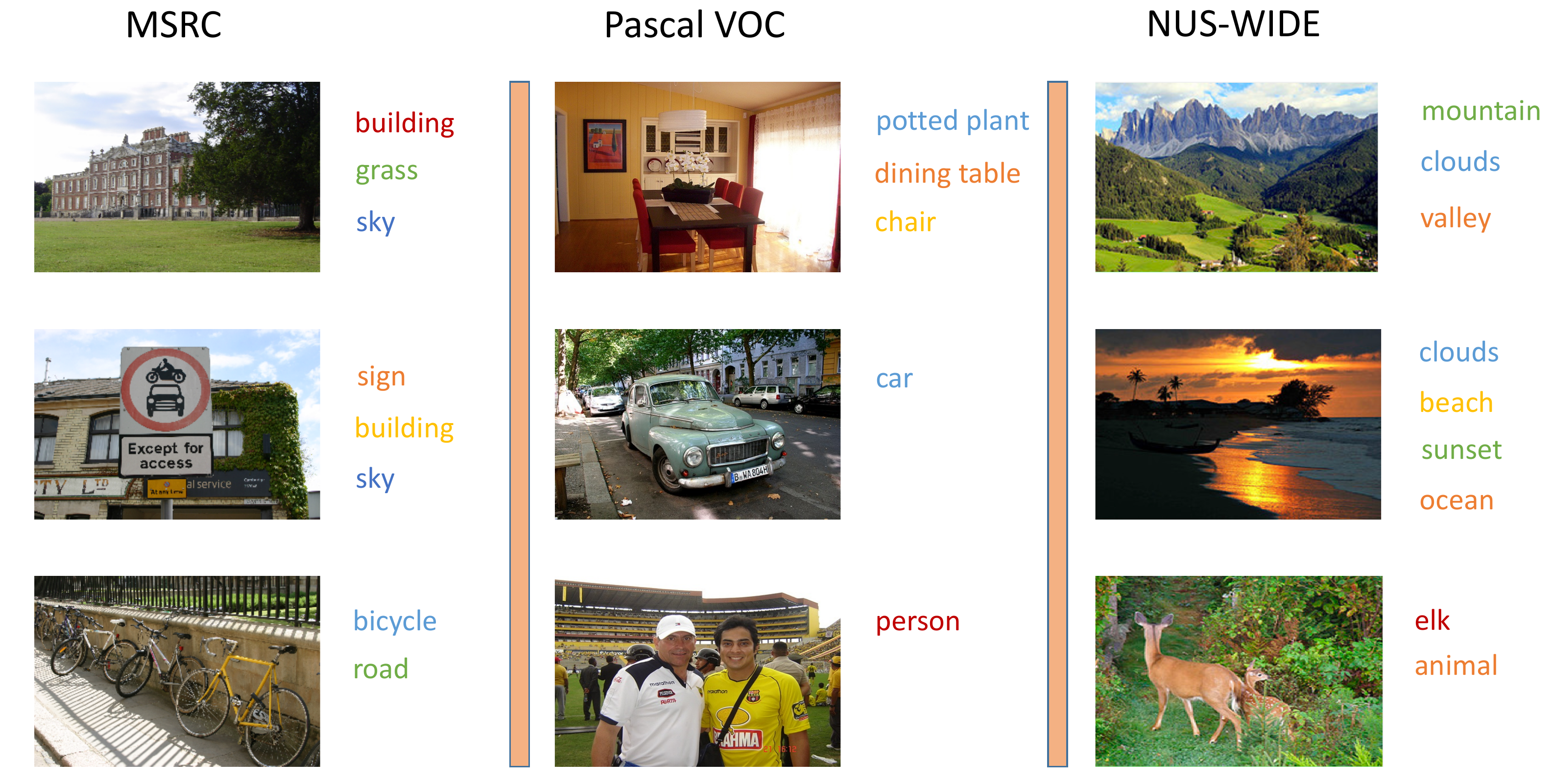}
\caption{Examplar images with corresponding labels from the MSRC, Pascal VOC and NUS-WIDE datasets respectively.}
\label{fig:example}
\end{figure*}

\section{Experiments}

In this section, we demonstrate the effectiveness of the proposed algorithm on six benchmark multi-label datasets including: \textit{Scene}~\cite{Boutell:PR2004} with $6$ scene classes such as mountain, beach and field; \textit{Emotions}~\cite{Trohidis:Emotions2008} with $6$ music emotion labels; \textit{Yeast}~\cite{Elisseeff:RankSVM2001} with $14$ gene functional categories (e.g., metabolism, energy); \textit{Mediamill}~\cite{Snoek:Mediamill2006} with $101$ semantic concept labels (e.g., military, desert, and basketball); \textit{MSRC} \cite{Winn:MSRC2005} with $23$ image class labels and \textit{SUNattribute} \cite{Patterson:SUN2014} with $102$ image class labels. 
 
 The data for the first four datasets can be directly downloaded from Mulan website\footnote{http://mulan.sourceforge.net/datasets-mlc.html}, where the corresponding data features have been extracted. We also follow the training and test subsets provided with the releases of the four datasets. MSRC\footnote{http://research.microsoft.com/en-us/projects/ObjectClassRecognition} is a Microsoft research labeled image dataset with $591$ images. Each image is represented by $512$ bag-of-words features on sampled patches. The SUNattribute dataset\footnote{https://cs.brown.edu/gen/sunattributes.html} contains $14,240$ images with $512$ GIST features. For MSRC and SUNattribute datasets, $10\%$ of the data from each class are used for training, while the rest $90\%$ are for testing. All experiments are repeated over 10 random training/test splits, the average results are reported.

The statistical information of these datasets used for experiments is summarized in Table~\ref{tab:dataset statistics}. From Table~\ref{tab:dataset statistics} we can observe that both label set sparsity (indicated by cardinality) and training data sparsity (estimated by $n/2^K$) are significant in these datasets. For example, each instance in the \textit{Scene} dataset only has an average of $1.074$ labels out of all 6 candidate labels. And each set of labels for the \textit{Mediamill} dataset is only occupied by as few as $0.04\%$ instances of the total number of training data.

Experimental results show that the proposed approach outperforms state-of-the-art methods on all six datasets and manifests strong generalization ability across different types of labels. Analysis of the transformed latent code demonstrates that our approach can effectively preserve the distribution of the original label vectors while alleviating the sparsity problem.

\subsection{Compared Methods and Evaluation Metrics}

\textbf{Compared Methods}. To validate the performance of our proposed DLST, we compare it with the following representative and related multi-label learning algorithms:

\begin{itemize} 
\item \textbf{BR}~\cite{Tsoumakas:BR2010}: Binary Relevance.  
\item \textbf{CPLST}~\cite{Chen:CPLST2012}: Conditional Principal Label Space Transformation.
\item \textbf{FAIE}~\cite{Lin:FAIE2014}: Feature-aware Implicit Label Space Encoding.
\item \textbf{MLLOC}~\cite{Huang:MLLOC2012}: Multi-Label Learning using Local Correlation.
\item \textbf{MC}~\cite{Cabral:MC2015}: Matrix Completion.
\item \textbf{MIML}~\cite{Vasisht:MIML2014}: Mutual Information for Multi-Label Classification.
\item \textbf{SLRM}~\cite{Jing:SLRM2015}: Semi-supervised Low-Rank Mapping.
\item \textbf{MRV}~\cite{Zhao:MRV2015}: Manifold Regularized Vector-valued Multi-Label Learning.
\item \textbf{LEML}~\cite{Yu:LEML2014}: Large Scale Empirical Risk Minimization Method with Missing Labels.
\end{itemize}

 BR is the baseline method, where each label is treated as an independent binary classification problem. CPLST, FAIE and MLLOC are label space reduction methods, which only use labeled instances as training set. MC, MIML and SLRM utilize both labeled and unlabeled instances for training. Meanwhile, SLRM, MRV and LEML are specifically designed for multi-label learning with missing labels. In the experiment, we adopt LibSVM~\cite{Chang:LibSVM2011} as the binary classifier for BR. In the learning stage, both CPLST and FAIE are coupled with linear regression for multi-label prediction. Unless otherwise specified, we set the parameters of the comparing methods according to what the authors supposed in the original papers or codes. As for DLST, a 10-fold cross-validation is performed by varying  $\lambda$ from $10^{-3}$ to $10^3$ with a stepsize of $10$. Results show that DLST yields stable performance around $\lambda=0.01$, which is used for DLST in the following experiments.
 
\textbf{Evaluation Metrics}. Performance evaluation for multi-label classification can be complicated since each instance is associated with a set of labels rather than a single one. Various metrics have been proposed based on the prediction likelihood with respect to each label, among which we adopt three widely-used evaluation metrics \textit{Average Precision}, \textit{Micro F1} and \textit{Macro F1} to quantitatively compare the performance of these multi-label classification methods. 

\textit{Average Precision (AP)} evaluates the average fraction of relevant labels ranked ahead of a particular label. The larger the value of AP, the better the performance. Its formal definition can be found in~\cite{Zhou:Metric2016}. 

\textit{Micro F1} and \textit{Macro F1} evaluate the micro average and macro average of the harmonic mean of precision and recall, respectively. As microaveraging and macroaveraging require binary indicator vectors, we consider the labels corresponding to the $r$ largest entries of the predicted vector as the predicted labels of each instance, where $r$ is set to be the average number of labels per instance. Therefore, from Table~\ref{tab:dataset statistics}, $r$ for Scene, Emotions, Yeast, Mediamill, MSRC and SUNattribute is $2, 2, 5, 5, 3$ and $16$ respectively. The bigger the value of \textit{Micro F1} and \textit{Macro F1}, the better the performance. Their formal definitions can be found in~\cite{Tsoumakas:RAKEL2007}.

\begin{table*}[t]
\centering
\begin{tabular}{l|a|b|a|b|a|b|a|b|a|b}
\hline \hline
Dataset & BR & CPLST & FAIE & MLLOC & MC & MIML & SLRM & MRV & LEML & DLST \\ \hline
\multicolumn{11}{c}{Average Precision $\uparrow$} \\ \hline
Scene
 & 0.4306 & 0.4492 & 0.4501 & 0.4327 & 0.4580 & 0.4888 & 0.5082 & 0.5125 & 0.4718 & \textbf{0.5365} \\
 \hline
Emotions
& 0.2734 & 0.2958 & 0.3012 & 0.3146 & 0.3068 & 0.3235 & 0.3487 & 0.3574 & 0.3128 & \textbf{0.3864} \\
\hline
Yeast 
& 0.3225 & 0.3364 & 0.3425 & 0.3389 & 0.3567 & 0.3842 & 0.4005 & 0.4082 & 0.3620 & \textbf{0.4312} \\
\hline
Mediamill
& 0.4086 & 0.4264 & 0.4265 & 0.4326 & 0.4509 & 0.4465 & 0.4691 & 0.4653 & 0.4324 & \textbf{0.4980} \\
\hline
MSRC
& 0.3145 & 0.3281 & 0.3346 & 0.2070 & 0.2353 & 0.2801 & 0.3864 & 0.3725 & 0.3376 & \textbf{0.4016} \\
\hline
SUNattribute
& 0.2876 & 0.3009 & 0.3387 & 0.2876 & 0.3052 & 0.2954 & 0.3286 & 0.3124 & 0.3092 & \textbf{0.3584} \\ 
\hline
\multicolumn{11}{c}{Micro F1 $\uparrow$} \\ \hline
Scene
 & 0.5987 & 0.6496 & 0.6528 & 0.6630 & 0.6282 & 0.6713 & 0.7022 & 0.7324 & 0.6825 & \textbf{0.7642} \\
 \hline
Emotions
& 0.3450 & 0.3633 & 0.3428 & 0.3596 & 0.3088 & 0.3694 & 0.4012 & 0.5424 & 0.5280 & \textbf{0.5642} \\
\hline
Yeast 
& 0.4435 & 0.4520 & 0.4631 & 0.4552 & 0.4328 & 0.4784 & 0.4950 & 0.6413 & 0.6086 & \textbf{0.6971} \\
\hline
Mediamill
& 0.4234 & 0.5785 & 0.6422 & 0.6381 & 0.6273 & 0.6412 & 0.6476 & 0.5283 & 0.5562 & \textbf{0.6632} \\
\hline
MSRC
& 0.4383 & 0.5109 & 0.5357 & 0.3692 & 0.4196 & 0.5538 & 0.5890 & 0.5726 & 0.3981 & \textbf{0.6235} \\
\hline
SUNattribute
& 0.4425 & 0.4605 & 0.4936 & 0.4441 & 0.4670 & 0.4521 & 0.5043 & 0.4631 & 0.4430 & \textbf{0.4926} \\ 
\hline
\multicolumn{11}{c}{Macro F1 $\uparrow$} \\ \hline
Scene
& 0.3153 & 0.3264 & 0.3358 & 0.3125 & 0.3562 & 0.3458 & 0.3745 & 0.3964 & 0.3692 & \textbf{0.4235} \\
 \hline
Emotions
& 0.1928 & 0.2034 & 0.2135 & 0.2234 & 0.2580 & 0.2542 & 0.2718 & 0.2826 & 0.2984 & \textbf{0.3260} \\
\hline
Yeast 
& 0.2436 & 0.2580 & 0.2624 & 0.2578 & 0.2842 & 0.2673 & 0.3016 & 0.3245 & 0.3326 & \textbf{0.3794} \\
\hline
Mediamill
& 0.1150 & 0.0982 & 0.1302 & 0.1399 & 0.1269 & 0.1298 & 0.1413 & 0.1526 & 0.1254 & \textbf{0.1738} \\
\hline
MSRC
& 0.3562 & 0.3317 & 0.3467 & 0.1048 & 0.2541 & 0.4083 & 0.4481 & 0.4468 & 0.3575 &  \textbf{0.4738} \\
\hline
SUNattribute
& 0.1842 & 0.2196 & 0.2630 & 0.1923 & 0.2507 & 0.2852 & 0.2687 & 0.2716 & 0.2535 & \textbf{0.3283} \\ 
\hline
\end{tabular}
\caption{Performance comparison for multi-label learning approaches on six datasets under different evaluation metrics. $\uparrow$ means the bigger the value, the better the performance.}
\label{tab:evaluation}
\end{table*}

\subsection{Experimental Results}
Quantitative results on all six datasets under three evaluation metrics are presented in Table~\ref{tab:evaluation}. From Table~\ref{tab:evaluation}, we can see that the proposed DLST performs better than or comparable to the other nine state-of-the-arts and baseline methods across all 18 configurations (6 datasets $\times$ 3 evaluation metrics). The superior performance of DLST across all three evaluation measures justifies our motivation of exploiting label distribution preservation during label transformation. In the following, we present a more detailed comparison between DLST and the other three categories of multi-label learning methods.

DLST outperforms label space reduction methods (CPLST, FAIE, MLLOC) by as much as $20\%, 28\%, 28\%, 17\%, 20\%, 19\%$ on the six datasets measured by average precision. This advantage demonstrates that DLST learns a higher quality latent code than the other three baselines in terms of approximating the original label space. Moreover, the latent space learned by DLST improves the original label space by revealing the comprehensive label correlations, thus alleviating the sparsity problem presented in multi-label classification. 
 
Moreover, DLST shows better performance than semi-supervised multi-label classification methods (MC, MIML, SLRM). The three semi-supervised baselines utilize abundant unlabeled data in the training process, which is believed to be able to boost the performance. However, these methods require a large number of training data to perform well. In contrast, DLST demonstrates comparable or even better performance with only 10\% of the training data used by the comparing semi-supervised baselines. The performance gain can be explained by the distribution employed in DLST, which can estimate and fulfill the comprehensive label correlations given only limited number of labeled training data.

Compared with baselines specifically targeting missing labels (SLRM, MRV, LEML), DLST almost always outperforms them by $14\%, 23\%, 19\%, 15\%, 19\%, 16\%$ on the six datasets with respect to average precision criterion. These results corroborate the effectiveness of DLST in exploiting inter-label correlations. In subsection~\ref{subsection:sparsity}, we will further compare the performance of these methods under varying number of labels for each instance.

\begin{table}[b]
\centering
\begin{tabular}{l|c|c|c|c|r}
\hline \hline
Dataset & type & n & d & K & card  \\ \hline
\rowcolor{Orange}
Pascal VOC & image & 9,963 & 2,048 & 20 & 1.560 \\
 \hline
NUS-WIDE & image & 269,648 & 500 & 81 & 1.869 \\
 \hline
\end{tabular}
\caption{Statistics of two large scale datasets used in the experiments.}
\label{tab:large scale dataset statistics}
\end{table}

\subsection{The Benefit of Latent Code of DLST}
In this subsection, to further study the superiority of DLST in learning a dense latent code, we conduct another set of experiments on two variants of DLST: {DLST}$_1$ and {DLST}$_2$. {DLST}$_1$ directly learns a regression function from the feature space to the original label space, while {DLST}$_2$ predicts the original label of test instances based on ML-KNN using the feature vector of training instances. The performance of these methods are evaluated on three datasets: Scene, Emotions and Yeast. Similar to previous experimental settings, the training and test subsets provided along with each dataset is adopted. The mean value and stand deviation of DLST and its two variants under the three evaluation metrics are recorded in Table~\ref{tab:variants}. 

From Tabel~\ref{tab:variants}, we can see that DLST outperforms the two variants by $20\%$ and $18\%$ respectively on the Scene dataset under average precision. This result verifies that sparsity in the original label space significantly deteriorates multi-label learning performance. It also suggests that the dense latent code learned by DLST is more effective in capturing inter-label correlations and more informative in predicting labels for multi-label instances.

\begin{figure}[t]
\centering
\includegraphics[width=\linewidth]{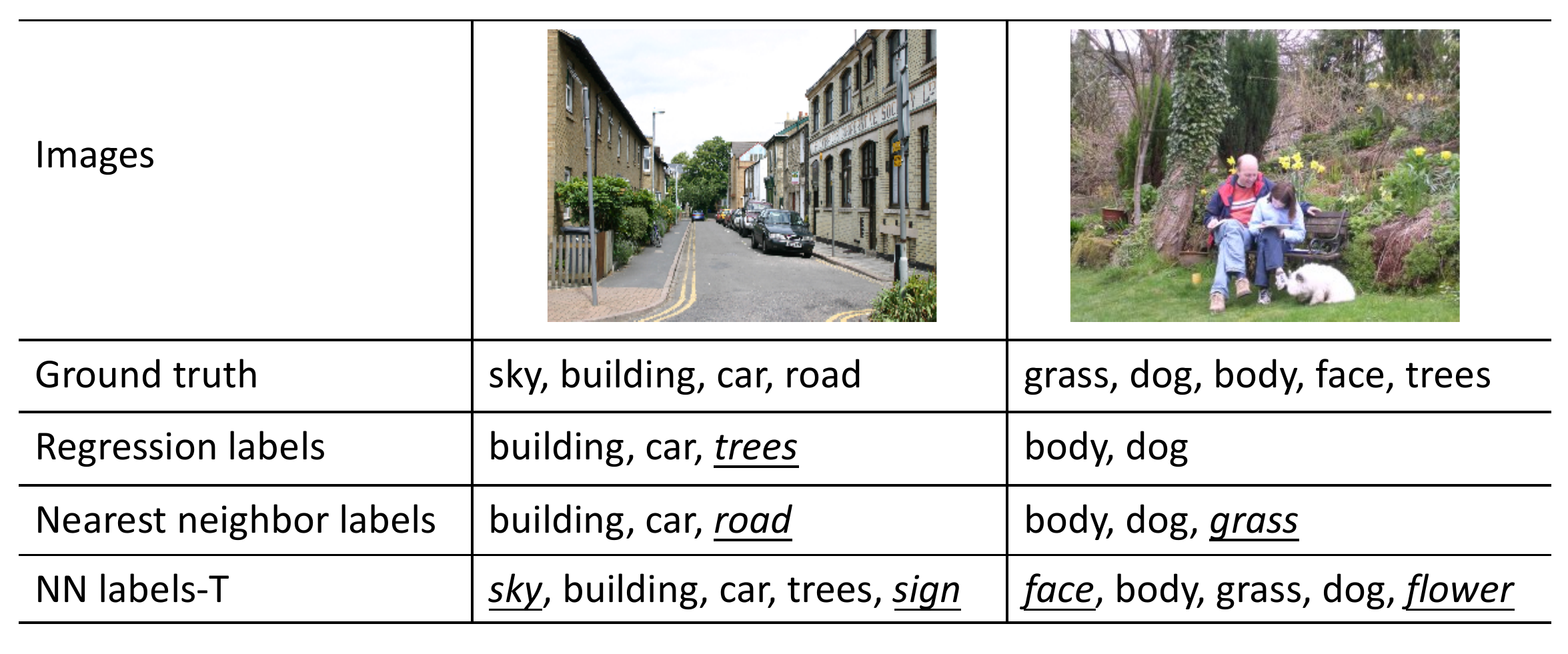}
\caption{Regression labels and nearest neighbor labels obtained using methods DLST$_1$, DLST$_2$ and DLST for images on MSRC dataset.}
\label{fig:labels}
\end{figure}

Figure~\ref{fig:labels} shows the regression labels and nearest neighbor labels for images on MSRC dataset. Regression labels are produced by applying DLST$_1$, while nearest neighbor labels are obtained by utilizing DLST$_2$. The difference between Nearest Neighbor labels and NN labels-T lies in the space where nearest neighbor searching takes place, specifically, the former occurs in the original feature space, while the latter occurs in the transformed latent space. From Figure~\ref{fig:labels}, it can be observed that DLST produces the most comprehensive label sets for multi-label images. The rationality lies in that the latent space derived by DLST captures the whole distribution of relative distances of any label pairs. Thus the probability of co-occurrence between labels can be more delicately predicted. 

\begin{table}
\centering
\begin{tabular}{l|c|c|c}
\hline \hline
Methods & Scene & Emotions & Yeast \\ \hline
\multicolumn{4}{c}{Average Precision $\uparrow$} \\ \hline
\rowcolor{Orange}
DLST$_1$
& 0.446 $\pm$ 0.005 & 0.297 $\pm$ 0.006 & 0.334 $\pm 0.002$  \\
 \hline
DLST$_2$
& 0.453 $\pm$ 0.020 & 0.304 $\pm$ 0.013 & 0.352 $\pm$ 0.009 \\
\hline
\rowcolor{Orange}
DLST
& \textbf{0.536 $\pm$ 0.001} & \textbf{0.386 $\pm$ 0.002} & \textbf{0.431 $\pm$ 0.004} \\
\hline
\multicolumn{4}{c}{Micro F1 $\uparrow$} \\ \hline
\rowcolor{Orange}
DLST$_1$
& 0.661 $\pm$ 0.008 & 0.465 $\pm$ 0.006 & 0.525 $\pm$ 0.003 \\
 \hline
DLST$_2$
& 0.690 $\pm$ 0.016 & 0.480 $\pm$ 0.014 & 0.594 $\pm$ 0.008 \\
\hline
\rowcolor{Orange}
DLST
& \textbf{0.763 $\pm$ 0.004} & \textbf{0.564 $\pm$ 0.003} & \textbf{0.697 $\pm$ 0.002} \\
\hline
\multicolumn{4}{c}{Macro F1 $\uparrow$} \\ \hline
\rowcolor{Orange}
DLST$_1$
& 0.306 $\pm$ 0.004 & 0.204 $\pm$ 0.008 & 0.253 $\pm$ 0.005 \\
 \hline
DLST$_2$
& 0.314 $\pm$ 0.012 & 0.218 $\pm$ 0.005 & 0.264 $\pm$ 0.008 \\
\hline
\rowcolor{Orange}
DLST
& \textbf{0.423 $\pm$ 0.002} & \textbf{0.326 $\pm$ 0.001} & \textbf{0.379 $\pm$ 0.003} \\
\hline
\end{tabular}
\caption{Experimental results (mean$\pm$std) of DLST and its two variants on Scene, Emotions and Yeast datasets across three evaluation metrics.}
\label{tab:variants}
\end{table}

\begin{figure}[t]
\centering
\includegraphics[width=\linewidth]{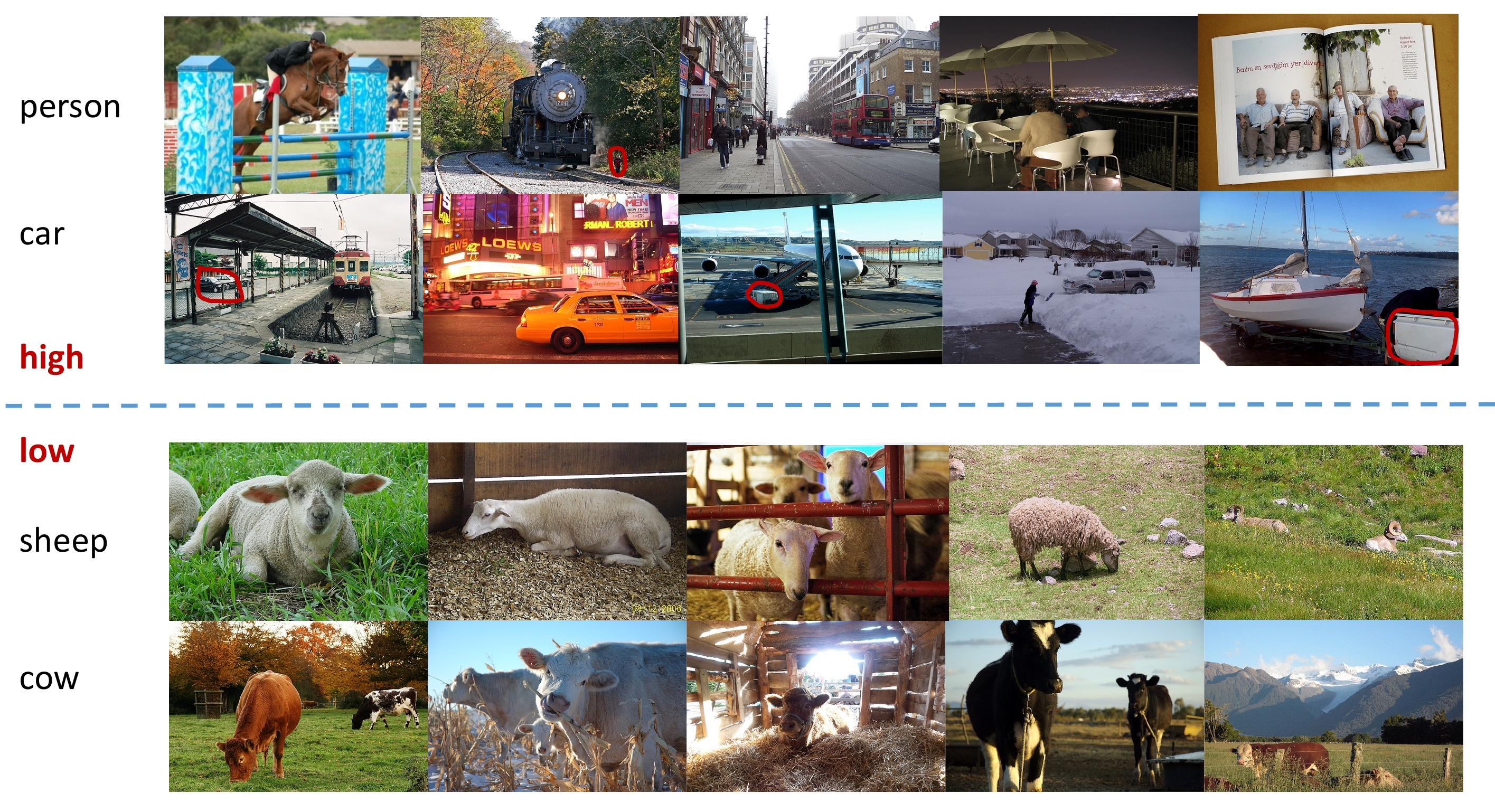}
\caption{Examplar images from classes that has (top) highest per-class-precision and (bottom) lowest per-class-precision on Pascal VOC dataset.}
\label{fig:largescale}
\end{figure}

\begin{figure*}[t]
\centering
\includegraphics[width=\linewidth]{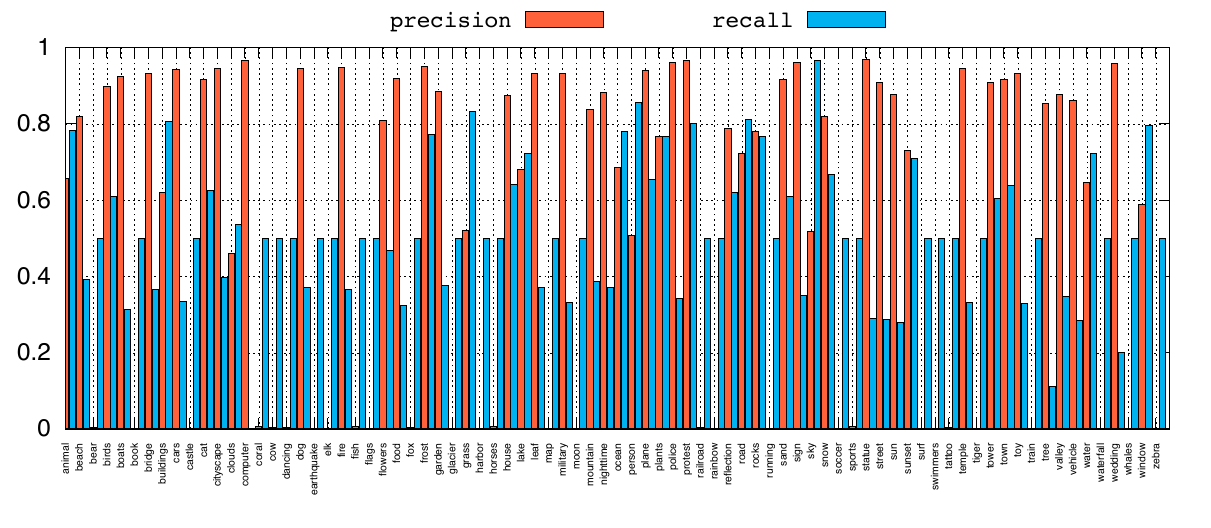}
\caption{The per-class precision and recall of DLST on NUS-WIDE dataset.}
\label{fig:NUSWIDE}
\end{figure*}

\begin{table*}
\centering
\begin{tabular}{l|p{.3cm}p{.4cm}p{.3cm}p{.3cm}p{.3cm}p{.3cm}p{.4cm}p{.3cm}p{.4cm}p{.3cm}p{.3cm}p{.3cm}p{.3cm}p{.4cm}p{.4cm}p{.3cm}p{.3cm}p{.3cm}p{.3cm}p{.3cm}|c}
\hline \hline
Methods & plane & bicycle & bird & boat & bottle & bus & car & cat & chair & cow & table & dog & horse & motor & person & plant & sheep & sofa & train & tv & mAP \\
\hline
$\#$labels & 1.2 & 1.9 & \underline{\textit{1.1}} & 1.4 & 2.4 & 2.0 & 1.7 & 1.4 & 2.5 & 1.4 & \underline{\textit{2.8}} & 1.6 & 1.9 & 1.9 & 2.0 & 2.3 & 1.3 & 2.4 & 1.3 & 2.2 & - \\
\rowcolor{Orange}
$\#$samples & 445 & 505 & 622 & 364 & 502 & 380 & 1536 & 676 & 1117 & 273 & 510 & 863 & 573 & 482 & \underline{\textit{4192}} & 527 & \underline{\textit{195}} & 727 & 522 & 534 & - \\
\hline
HCP-1000C & 95.1 & 90.1 & 92.8 & 89.9 & 51.5 & 80.0 & 91.7 & 91.6 & 57.7 & 77.8 & 70.9 & 89.3 & 89.3 & 85.2 & 93.0 & 64.0 & 85.7 & 62.7 & 94.4 & 78.3 & 81.5 \\
\rowcolor{Orange}
CNN-RNN & 96.7 & 83.1 & 94.2 & 92.8 & 61.2 & 82.1 & 89.1 & 94.2 & 64.2 & 83.6 & 70.0 & 92.4 & 91.7 & 84.2 & 93.7 & 59.8 & 93.2 & 75.3 & 99.7 & 78.6 & 84.0 \\
DLST$_1$ & 95.5 & 93.1 & 92.4 & 91.8 & 90.2 & 83.4 & 73.6 & 85.4 & 67.5 & 85.3 & 84.2 & 80.4 & 84.6 & 84.2 & 40.8 & 85.2 & 84.7 & 83.2 & 82.8 & 74.6 & 79.6 \\
\hline
\rowcolor{Orange}
DLST & \textbf{96.8} & \textbf{95.2} & \textbf{93.4} & \textbf{96.3} & \textbf{95.1} & \textbf{96.4} & \textbf{84.2} & \textbf{93.3} & \textbf{88.4} & \textbf{97.2} & \textbf{94.8} & \textbf{91.8} & \textbf{94.3} & \textbf{94.8} & \textit{\textbf{51.5}} & \textbf{94.8} & \textit{\textbf{98.2}} & \textbf{93.0} & \textbf{94.2} & \textbf{94.6} & \textbf{85.7} \\
\hline
\end{tabular}
\caption{The per-class precision and mAP of DLST and compared methods on Pascal VOC dataset. The biggest and smallest number of labels-per-instance as well as samples-per-class are labeled in italic and underlined. The highest and lowest average precision-per-class are shown in boldface and italic.}
\label{tab:Pascal}
\end{table*}

\subsection{Large Scale Datasets}
To evaluate the performance of DLST on large scale datasets, we additionally employ two multi-label datasets: Pascal VOC2007~\cite{Everingham:PASCAL2010} and NUS-WIDE~\cite{Chua:NUS-WIDE2009}. 

Pascal Visual Object Classes Challenge (VOC) datasets have been widely used as the benchmark for multi-label classification. VOC 2007 dataset contains $9,963$ images with $20$ labels. Each image in this dataset is represented by $2,048$-dimensional deep CNN feature, which is  generated by ResNet-50~\cite{He:CVPR2016} pretrained on ImageNet database. Images in the \textit{train} and \textit{validation} subsets are used as training data, while images in the \textit{test} subset are utilized as testing data. Therefore, the training/test split adopted in the experiment are $5011/4592$ images.

NUS-WIDE dataset is a web image dataset, which contains $269,648$ images and $5,018$ tags collected from Flickr. There are $1,000$ tags after removing noisy and rare tags. These images are further manually annotated into 81 concept groups, e.g., \textit{sunset, clouds, beach, mountain, animal} as shown in Figure~\ref{fig:example}. The $500$ Bag-of-Words features based on SIFT descriptions are adopted to represent each image in this dataset. Among the images, $8,000$ images are utilized for training, and $8,000$ images are employed for testing. All experiments are conducted over 10 random training/test subsets of data, and the average performance are recorded. Table~\ref{tab:large scale dataset statistics} summarizes more detailed characteristics of these two datasets. Similar notations are adopted as in Table~\ref{tab:dataset statistics}.

Since other baselines cannot deal with such large amounts of data (out of memory), we compare the proposed method with two state-of-the-art deep learning methods \textbf{HCP-1000C}~\cite{Wei:HCP2014} and \textbf{CNN-RNN}~\cite{Wang:CNN-RNN2016} as well as the variant of our methods DLST$_1$. The precision and recall of predicted labels are employed as evaluation metrics. For each image, the precision records the number of correctly annotated labels divided by the number of generated labels; while the recall is defined as the number of correctly annotated labels divided by the number of ground-truth labels. We additionally compute the per-class precision and mAP for both datasets.

From Table~\ref{tab:Pascal}, we can observe that DLST achieves consistently higher average precision than other comparing methods across all label classes. Moreover, it is interesting to note that DLST performs best on class \textit{sheep} and worst on class \textit{person}, which correlates negatively with the number of training instances for each class. 

This can be explained from two aspects, on one hand, classes with more training samples tend to have noisy correlations with other labels, as shown in Figure~\ref{fig:largescale}, images for person and car are either occluded or dominated by other objects in the image. Therefore, the distances between a certain label and all the other labels are relatively the same, leading to approximately uniform distribution according to DLST. Therefore, the discriminative information that helps to identify a certain class is overwhelmed by the noise present in large volume of diverse training instances. On the other hand, classes with less training samples are likely to develop simple and clear relationships with a limited number of other labels, e.g., sheep and cow almost always relate to plant, whereas seldom relate to boat. Thus, according to the distribution proposed by DLST, these labels have a prominently higher probability to co-occur with a small group of specific labels, which dramatically reduces the difficulty of recognizing them in various images. Actually the relationship holds as long as relatively equal number of labels-per-instance are assigned each class, which can be well observed in Pascal VOC dataset (the first row in Table~\ref{tab:Pascal}). This observation further justifies the advantage of leveraging distribution to tackle sparsity in multi-label classification. 

Figure~\ref{fig:NUSWIDE} shows the per-class precision and recall of DLST on NUS-WIDE dataset. It can be seen that DLST achieves high precision and low recall on classes with few labels, Representative classes include \textit{computer, protest and wedding}. The reason is that DLST tends to stop predicting more labels for sparsely correlated labels. While on other classes such as \textit{map, book and rainbow}, which have larger label cardinality, DLST achieves low precision and high recall. This may be caused by the insufficient standard training data for these classes. There are also some classes that obtain comparable precision and recall, such as \textit{clouds, grass, person}. Notably, these concepts are ubiquitous among all the images in NUS-WIDE dataset. Moreover, the mean precision and recall of DLST averaged over all classes are $0.5375$ and $0.4489$ respectively, which is $32.6\%$ and $47.6\%$ higher than state-of-the-art CNN-RNN methods. However, it is worthy to note that this result is yielded by randomly selecting $8,000$ images for training and $8,000$ for testing, rather than employing the whole dataset for implementation.

\subsection{The Advantage of DLST in Tackling Sparsity}
\label{subsection:sparsity}

\subsubsection{Sparsity I: Label Set Sparsity for Instances}

\begin{figure}[h]
\centering
\begin{subfigure}{.48\linewidth}
  \includegraphics[width=\linewidth]{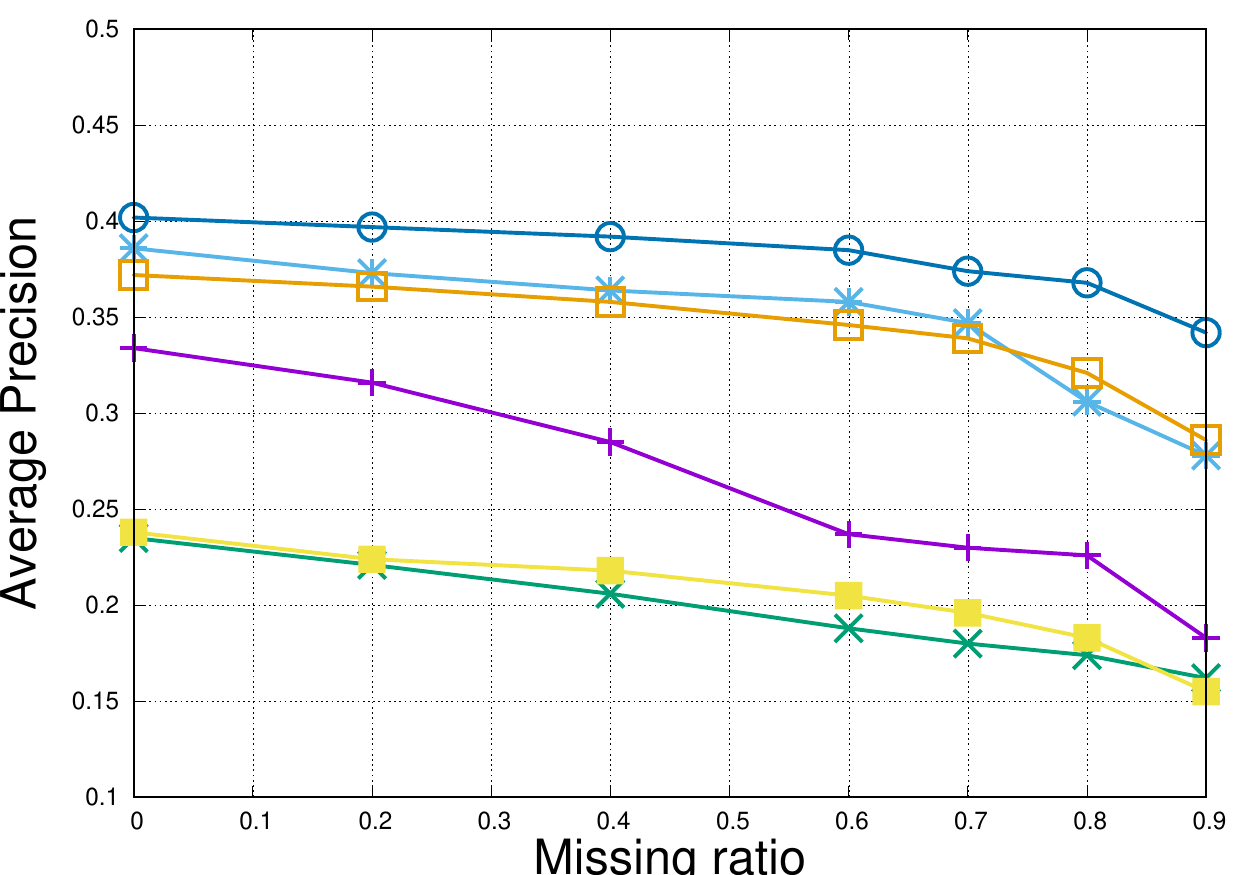}
  \caption{AP on MSRC}
  \label{fig:mrsc-AP}
\end{subfigure}
\begin{subfigure}{.48\linewidth}
  \includegraphics[width=\linewidth]{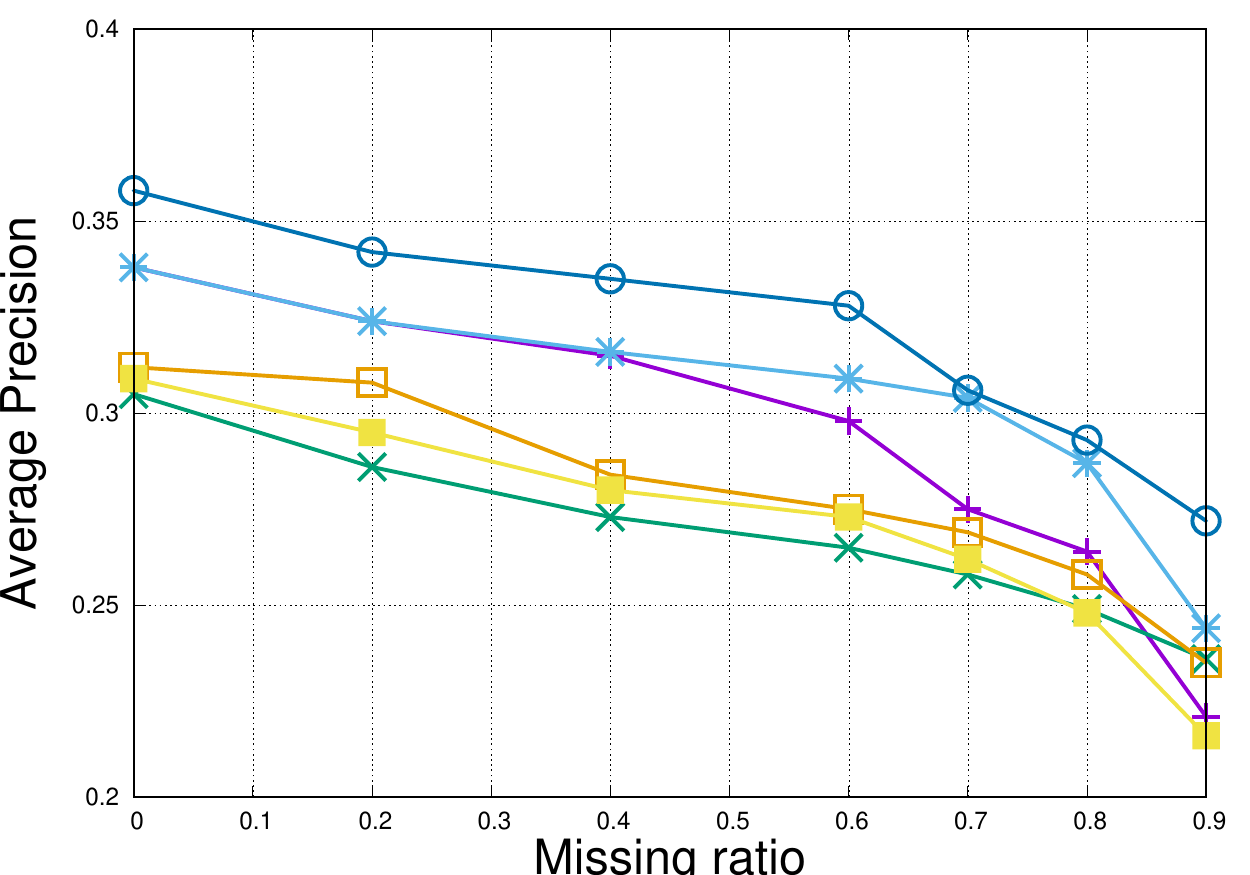}
  \caption{AP on SUN}
  \label{fig:SUN-AP}
\end{subfigure}
\begin{subfigure}{.48\linewidth}
  \includegraphics[width=\linewidth]{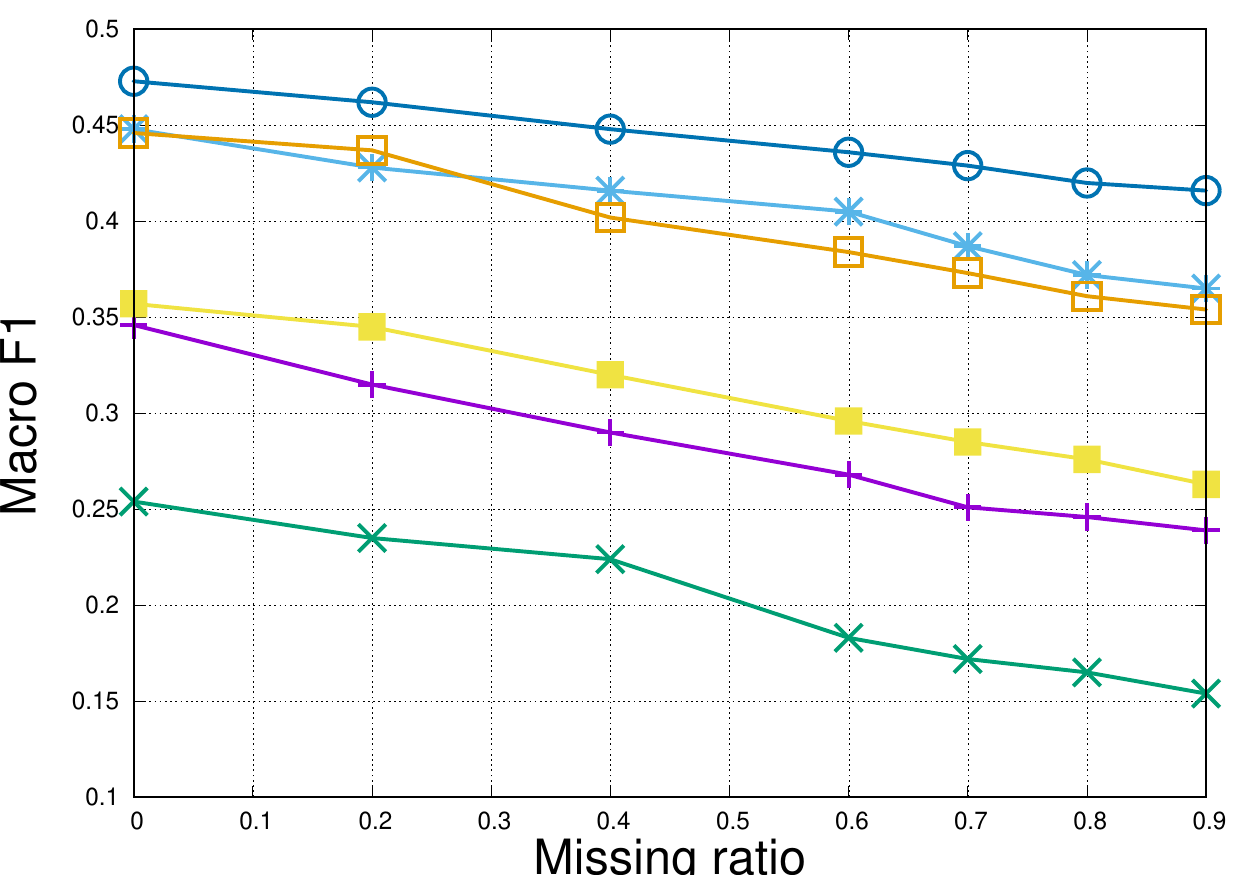}
  \caption{Macro F1 on MSRC}
  \label{fig:msrc-F1}
\end{subfigure}
\begin{subfigure}{.48\linewidth}
  \includegraphics[width=\linewidth]{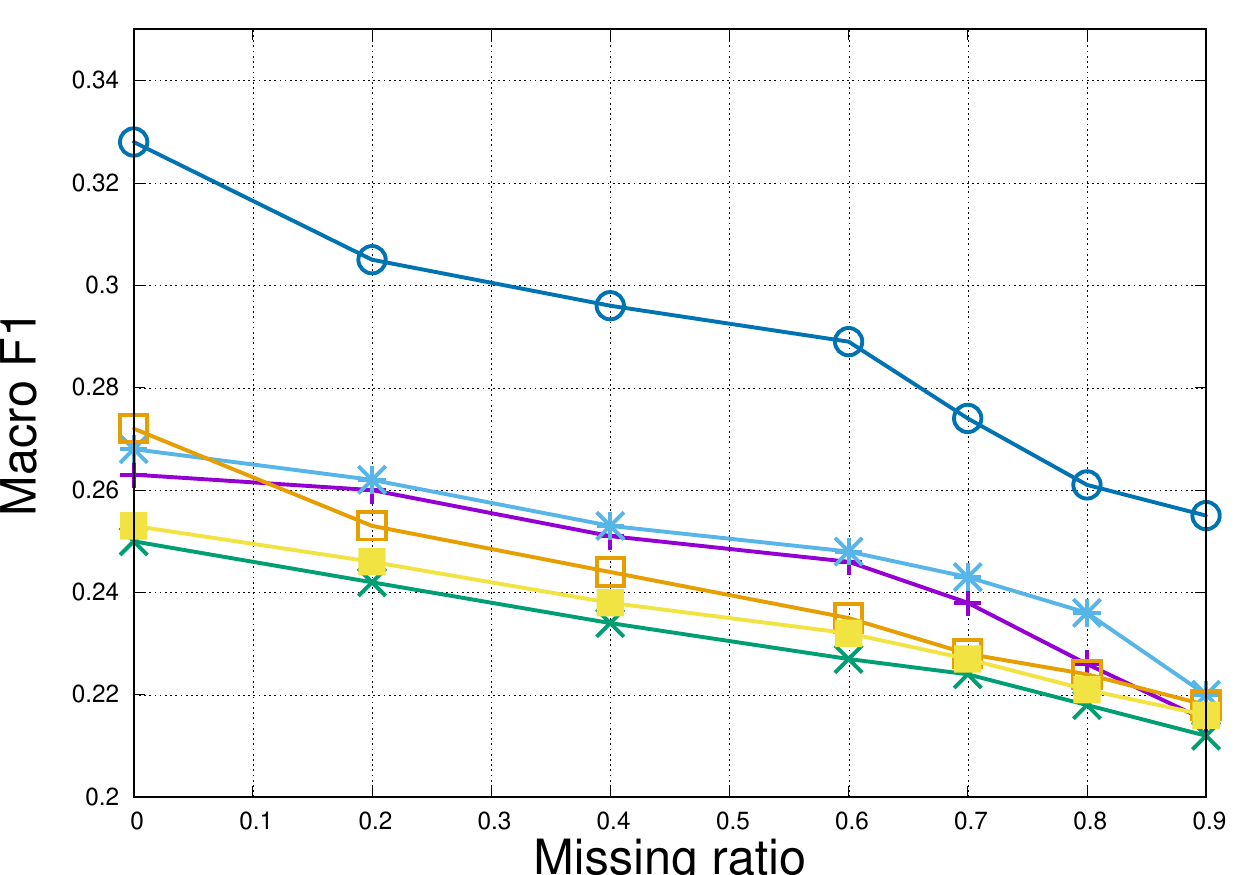}
  \caption{Macro F1 on SUN}
  \label{fig:SUN-F1}
\end{subfigure}
\par\bigskip
\begin{subfigure}{.9\linewidth}
  \includegraphics[width=\linewidth]{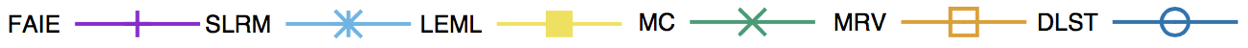}
  \label{fig:legend_1}
\end{subfigure}
\caption{Performance comparison of six methods with different proportions of missing labels on MSRC and SUN.}
\vspace{-2mm}
\label{fig:Missing ratio}
\end{figure}

We investigate the performance of DLST in handling label set sparsity by conducting experiments with missing labels. This setting also facilitates comparison with other baselines. The experimental data are generated on datasets MSRC and SUN with 10\% of the data from each class for training and the remaining for testing. The experiments are conducted with $\eta=\{20\%, 40\%, 60\%, 70\%, 80\%, 90\%\}$ missing labels on the labeled training instances. For each missing ratio, we randomly drop $\{20\%, 40\%, 60\%, 70\%, 80\%, 90\%\}$ of the observed labels. To avoid empty class or instances with no positive labels, at least one instance is kept for each class and at least one positive label is kept for each instance. Then the label vector $y_i$ for training instance $x_i$ is reset according to the protocol: $y_{ij}=1$ if the $i$-th instance is assigned the $j$-th label, and $y_{ij}=0$ otherwise. Note that $y_{ij}=0$ may indicate missing label or negative label. 

The proposed method is compared with FAIE, MC, SLRM, MRV, LEML, and the results are shown in Figure~\ref{fig:Missing ratio}. We can observe that DLST shows notable performance gain over other methods across all numbers of labels assigned to each instance. Also, the performance of DLST increases relatively less than the comparing methods as the average number of labels per instance increases. This is because the label correlations are more prominent with an increasing number of labels per instance, which benefits most methods significantly. However, DLST performs relatively stable and depends less on the extra given labels, since it can capture the comprehensive relationship between labels by using distribution alignment. As shown in Figure~\ref{fig:Missing ratio}, the minimum number of labels per instance for DLST to obtain sufficient information (i.e., performance variance less than $1\%$ in AP) is $3$ and $6$ for MSRC and SUN datasets respectively, which is much less than the comparing baselines.

\subsubsection{Sparsity II: Training Data Sparsity for Labels}

\begin{figure}[t]
\centering
\begin{subfigure}{.48\linewidth}
  \includegraphics[width=\linewidth]{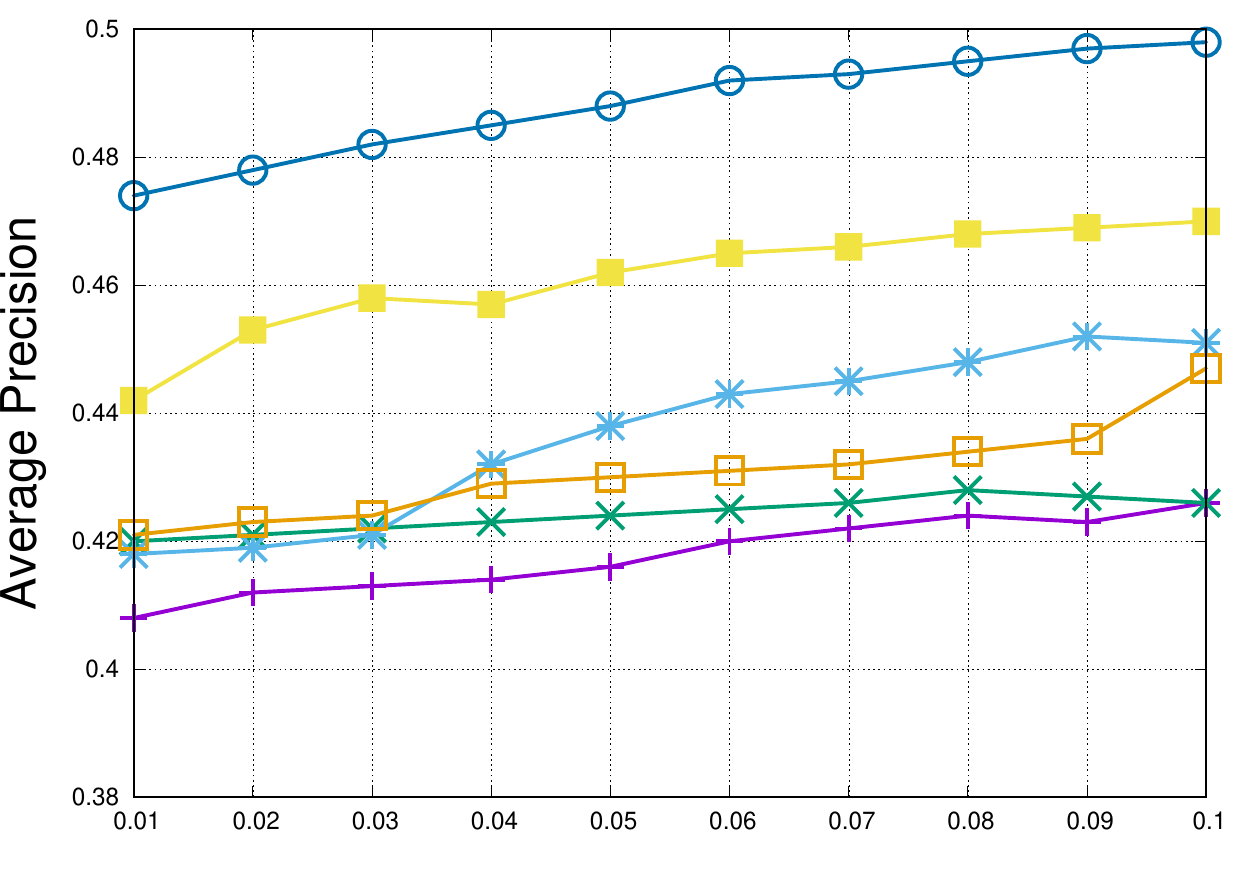}
  \caption{AP on Mediamill}
  \label{fig:Mediamill-AP}
\end{subfigure}
\begin{subfigure}{.48\linewidth}
  \includegraphics[width=\linewidth]{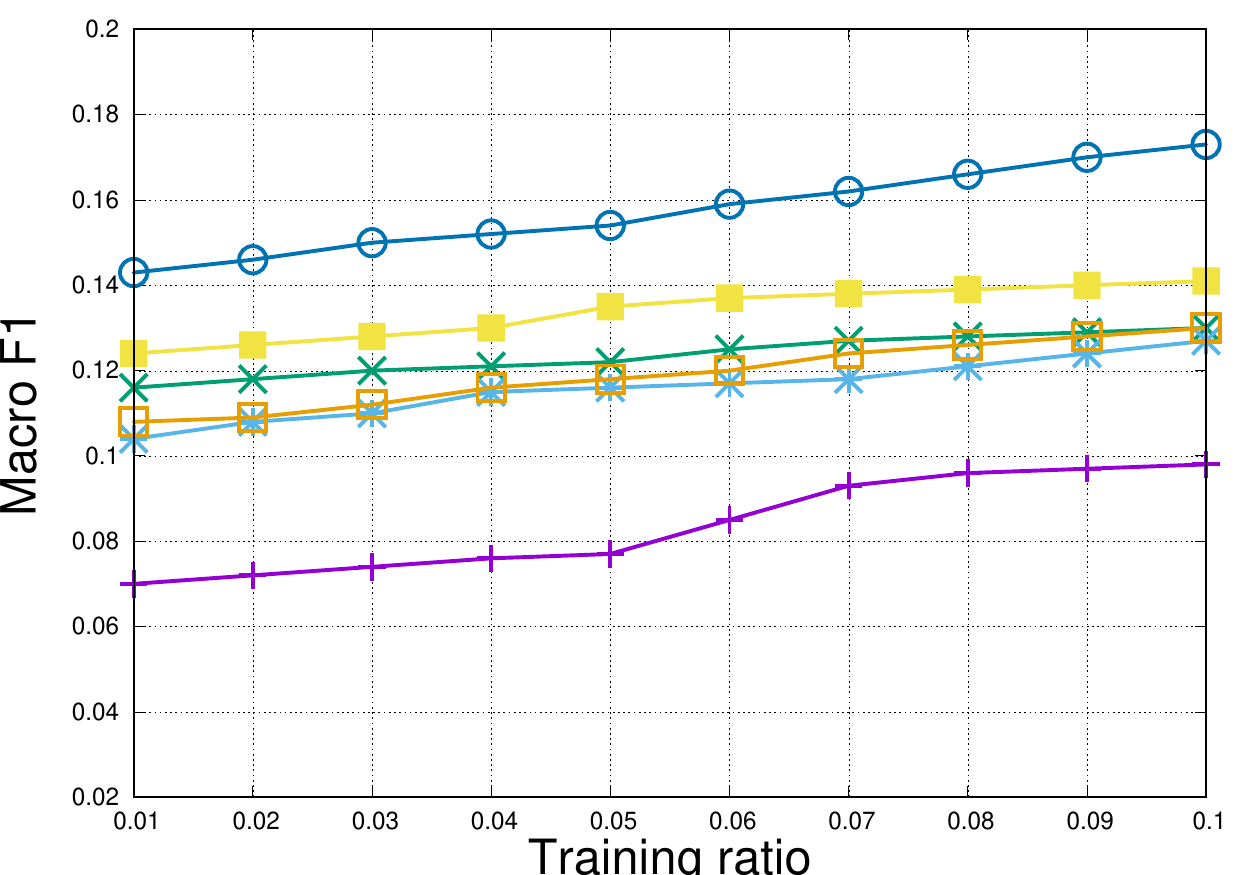}
  \caption{Macro F1 on Mediamill}
  \label{fig:Mediamill-AP}
\end{subfigure}
\par\bigskip
\begin{subfigure}{.9\linewidth}
  \includegraphics[width=\linewidth]{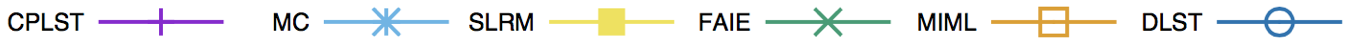}
  \label{fig:legend_1}
\end{subfigure}
\caption{Performance comparison of six methods with varying number of labeled training data on Mediamill.}
\vspace{-2mm}
\label{fig:Training ratio}
\end{figure}

We additionally investigate the effectiveness of DLST in tackling training data sparsity by varying the number of training instances for each class from $1\%$ to $10\%$ with a stepsize of $1\%$. For a given percentage, a corresponding number of training instances are randomly sampled for 10 times, and the resulting average precision are recorded. The experimental results of CPLST, MC, SLAM, FAIE, MIML are shown in Figure~\ref{fig:Training ratio}. Although the performance of all the methods degrade with a decreasing number of labeled training instances, DLST achieves a relatively stable performance across all training ratios and consistently outperforms comparing baselines. This can be explained by the capability of DLST to exploit inter-label correlations encoded in distribution. Given the fact that the comparing methods usually need $30\%$ training instances to saturate. In contrast, DLST only requires as few as $10\%$ training instances per class to gain the highest classification performance (with less than $1\%$ performance variance). This verifies that by using distribution, the label correlations can be more effectively and efficiently exploited, thus remarkably reducing the requirement on training size.

\subsection{Further Analysis}
\begin{figure}[h]
\centering
\includegraphics[width=\linewidth]{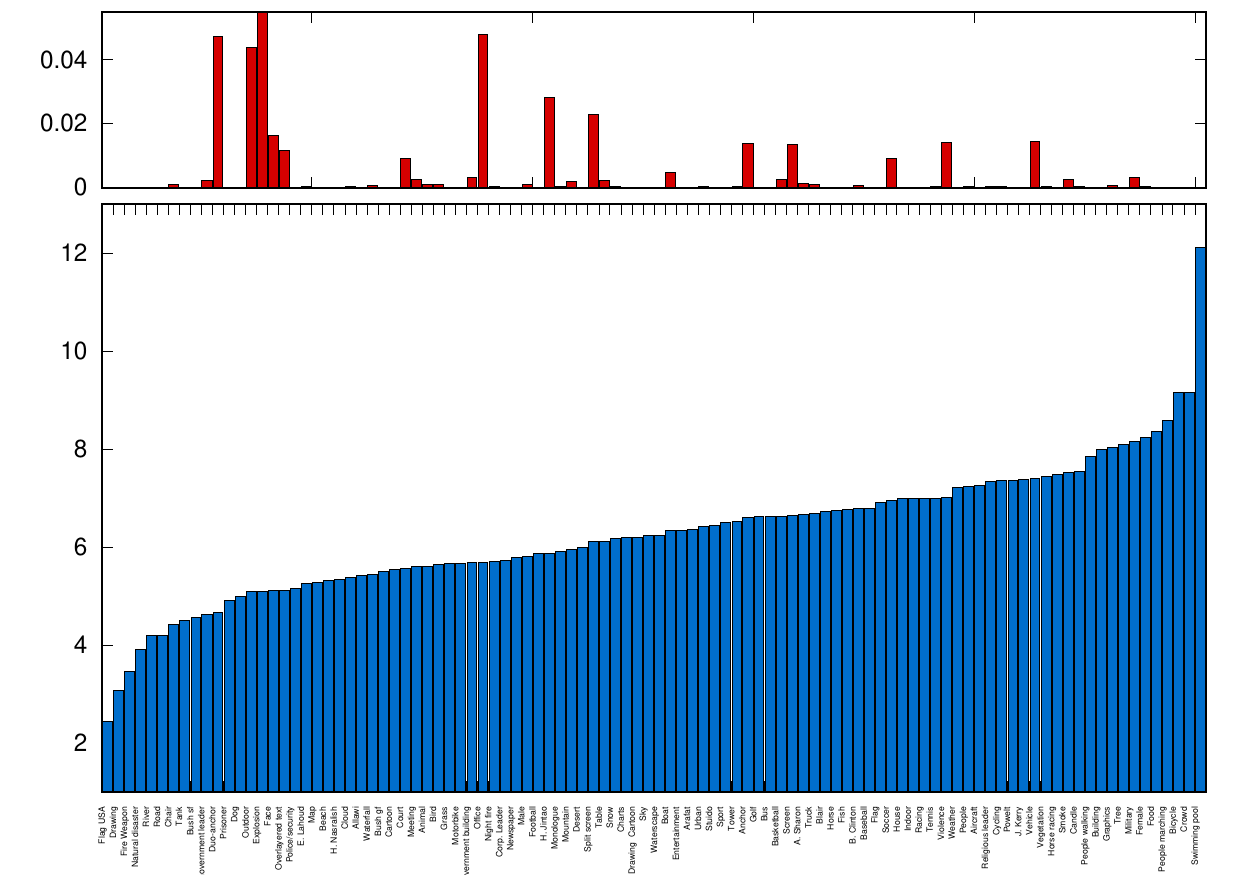}
\caption{Top: The average precision improvement for each class on the Mediamill dataset. Bottom: Average number of concurrent labels for true positive instances for each label class. All sorted according to number of concurrent labels in ascending order.}
\label{fig:AP_improvement}
\end{figure}

The effectiveness of our approach has been quantitatively evaluated in Table~\ref{tab:evaluation}. We further present the qualitative analysis of the learned latent code of DLST to illustrate its capability of revealing label correlations as well as addressing the sparsity problems. In Figure~\ref{fig:AP_improvement}, we show the average precision for each class on the Mediamill dataset.

To investigate the capability of DLST in tackling sparsity in the original label space, we study the Average Precision improvement for classes with different number of concurrent labels. From Figure~\ref{fig:AP_improvement} we can see that, AP improvement is more prominent for classes with small number of average concurrent labels. Typical examples include classes such as \textit{car}, \textit{grass}, \textit{beach} and \textit{waterscape}, where an AP improvement of $6\%, 5.8\%, 5.6\%, 5.5\%$ can be observed respectively. Although these labels are sparsely correlated to other labels for each instance, nonetheless, by learning from the distribution of all the possible labels, DLST is able to predict the corresponding labels with high precision. The result justifies our proposition that distribution reveals more comprehensive information about the label correlations which can be utilized to enhance multi-label classification performance.

\section{Conclusion}

To tackle the label sparsity and resolve label correlation for multi-label classification, in this paper, a distribution-based label space transformation method is proposed. By introducing the concept of distribution, more comprehensive relationship among labels of training instances can be captured. A much denser latent code is learned, enabling the proposed model to cope with both label set sparsity and training data sparsity where most multi-label classification methods fail to work effectively. The proposed model is especially successful in capturing a set of distinctive concurrent label patterns from a large pool of label vocabularies, which offers significant benefits to multi-label classification. Extensive experimental results demonstrate that DLST is superior to state-of-the-art multi-label learning methods under various percentage of labeled data and missing labels. 

\bibliographystyle{./IEEEtran}
\bibliography{./DLST}

% Generated by IEEEtranS.bst, version: 1.14 (2015/08/26)
\begin{thebibliography}{10}
\providecommand{\url}[1]{#1}
\csname url@samestyle\endcsname
\providecommand{\newblock}{\relax}
\providecommand{\bibinfo}[2]{#2}
\providecommand{\BIBentrySTDinterwordspacing}{\spaceskip=0pt\relax}
\providecommand{\BIBentryALTinterwordstretchfactor}{4}
\providecommand{\BIBentryALTinterwordspacing}{\spaceskip=\fontdimen2\font plus
\BIBentryALTinterwordstretchfactor\fontdimen3\font minus
  \fontdimen4\font\relax}
\providecommand{\BIBforeignlanguage}[2]{{%
\expandafter\ifx\csname l@#1\endcsname\relax
\typeout{** WARNING: IEEEtranS.bst: No hyphenation pattern has been}%
\typeout{** loaded for the language `#1'. Using the pattern for}%
\typeout{** the default language instead.}%
\else
\language=\csname l@#1\endcsname
\fi
#2}}
\providecommand{\BIBdecl}{\relax}
\BIBdecl

\bibitem{Agrawal:KeywordSuggestion2013}
R.~Agrawal, A.~Gupta, Y.~Prabhu, and M.~Varma, ``Multi-label learning with
  millions of labels: recommending advertiser bid phrases for web pages,'' in
  \emph{WWW}, 2013, pp. 13--24.

\bibitem{Barutcuoglu:Bio2006}
Z.~Barut{\c{c}}uoglu, R.~E. Schapire, and O.~G. Troyanskaya, ``Hierarchical
  multi-label prediction of gene function,'' \emph{Bioinformatics}, vol.~22,
  no.~7, pp. 830--836, 2006.

\bibitem{Bi:AAAI2014}
W.~Bi and J.~T. Kwok, ``Multilabel classification with label correlations and
  missing labels,'' in \emph{AAAI}, 2014, pp. 1680--1686.

\bibitem{Boutell:PR2004}
M.~R. Boutell, J.~Luo, X.~Shen, and C.~M. Brown, ``Learning multi-label scene
  classification,'' \emph{PR}, vol.~37, no.~9, pp. 1757--1771, 2004.

\bibitem{Bradley:ICML2010}
J.~K. Bradley and C.~Guestrin, ``Learning tree conditional random fields,'' in
  \emph{ICML}, 2010, pp. 127--134.

\bibitem{Cabral:MC2015}
R.~S. Cabral, F.~D. la~Torre, J.~P. Costeira, and A.~Bernardino, ``Matrix
  completion for weakly-supervised multi-label image classification,''
  \emph{TPAMI}, vol.~37, no.~1, pp. 121--135, 2015.

\bibitem{Chang:LibSVM2011}
C.~Chang and C.~Lin, ``{LIBSVM:} {A} library for support vector machines,''
  \emph{ACM TIST}, vol.~2, no.~3, pp. 27:1--27:27, 2011.

\bibitem{Chen:FastTag2013}
M.~Chen, A.~X. Zheng, and K.~Q. Weinberger, ``Fast image tagging,'' in
  \emph{ICML}, 2013, pp. 1274--1282.

\bibitem{Chen:CPLST2012}
Y.~Chen and H.~Lin, ``Feature-aware label space dimension reduction for
  multi-label classification,'' in \emph{NIPS}, 2012, pp. 1538--1546.

\bibitem{Cheng:IBLR2009}
W.~Cheng and E.~H{\"{u}}llermeier, ``Combining instance-based learning and
  logistic regression for multilabel classification,'' \emph{ML}, vol.~76, no.
  2-3, pp. 211--225, 2009.

\bibitem{Chow:TIT1968}
C.~K. Chow and C.~N. Liu, ``Approximating discrete probability distributions
  with dependence trees,'' \emph{{IEEE} Trans. Information Theory}, vol.~14,
  no.~3, pp. 462--467, 1968.

\bibitem{Chua:NUS-WIDE2009}
T.~Chua, J.~Tang, R.~Hong, H.~Li, Z.~Luo, and Y.~Zheng, ``{NUS-WIDE:} a
  real-world web image database from national university of singapore,'' in
  \emph{CIVR}.

\bibitem{Elisseeff:RankSVM2001}
A.~Elisseeff and J.~Weston, ``A kernel method for multi-labelled
  classification,'' in \emph{NIPS}, 2001, pp. 681--687.

\bibitem{Everingham:PASCAL2010}
M.~Everingham, L.~Gool, C.~K. Williams, J.~Winn, and A.~Zisserman, ``The pascal
  visual object classes (voc) challenge,'' \emph{Int. J. Comput. Vision},
  vol.~88, no.~2, pp. 303--338, 2010.

\bibitem{Funkranz:LabelRank2008}
J.~F{\"{u}}rnkranz, E.~H{\"{u}}llermeier, E.~Loza~Menc\'{i}a, and K.~Brinker,
  ``Multilabel classification via calibrated label ranking,'' \emph{ML},
  vol.~73, no.~2, pp. 133--153, 2008.

\bibitem{Hariharan:ICML2010}
B.~Hariharan, L.~Zelnik{-}Manor, S.~V.~N. Vishwanathan, and M.~Varma, ``Large
  scale max-margin multi-label classification with priors,'' in \emph{ICML},
  2010, pp. 423--430.

\bibitem{He:CVPR2016}
K.~He, X.~Zhang, S.~Ren, and J.~Sun, ``Deep residual learning for image
  recognition,'' in \emph{CVPR}, 2016, pp. 770--778.

\bibitem{Huang:MLLOC2012}
S.~Huang and Z.~Zhou, ``Multi-label learning by exploiting label correlations
  locally,'' in \emph{AAAI}, 2012.

\bibitem{Jiang:TPG2012}
J.~Jiang, ``Multi-label learning on tensor product graph,'' in \emph{AAAI},
  2012, pp. 956--962.

\bibitem{Jing:SLRM2015}
L.~Jing, L.~Yang, J.~Yu, and M.~K. Ng, ``Semi-supervised low-rank mapping
  learning for multi-label classification,'' in \emph{CVPR}, 2015, pp.
  1483--1491.

\bibitem{Kong:KDD2013}
X.~Kong, B.~Cao, and P.~S. Yu, ``Multi-label classification by mining label and
  instance correlations from heterogeneous information networks,'' in
  \emph{KDD}, 2013, pp. 614--622.

\bibitem{Kong:LabelSet2013}
X.~Kong, M.~K. Ng, and Z.~Zhou, ``Transductive multilabel learning via label
  set propagation,'' \emph{TKDE}, vol.~25, no.~3, pp. 704--719, 2013.

\bibitem{Li:CVPR2016}
Q.~Li, M.~Qiao, W.~Bian, and D.~Tao, ``Conditional graphical lasso for
  multi-label image classification,'' in \emph{CVPR}, 2016, pp. 2977--2986.

\bibitem{Li:UAI2014}
X.~Li, F.~Zhao, and Y.~Guo, ``Multi-label image classification with {A}
  probabilistic label enhancement model,'' in \emph{UAI}, 2014, pp. 430--439.

\bibitem{Lin:FAIE2014}
Z.~Lin, G.~Ding, M.~Hu, and J.~Wang, ``Multi-label classification via
  feature-aware implicit label space encoding,'' in \emph{ICML}, 2014, pp.
  325--333.

\bibitem{Liu:NIPS2015}
W.~Liu and I.~W. Tsang, ``On the optimality of classifier chain for multi-label
  classification,'' in \emph{NIPS}, 2015, pp. 712--720.

\bibitem{Patterson:SUN2014}
G.~Patterson, C.~Xu, H.~Su, and J.~Hays, ``The {SUN} attribute database: Beyond
  categories for deeper scene understanding,'' \emph{International Journal of
  Computer Vision}, vol. 108, no. 1-2, pp. 59--81, 2014.

\bibitem{Read:PPT2008}
J.~Read, ``A pruned problem transformation method for multi-label
  classification,'' in \emph{Proc. New Zealand Computer Science Research
  Student Conference}, 2008, pp. 143--150.

\bibitem{Read:CC2011}
J.~Read, B.~Pfahringer, G.~Holmes, and E.~Frank, ``Classifier chains for
  multi-label classification,'' \emph{ML}, vol.~85, no.~3, pp. 333--359, 2011.

\bibitem{Schapire:ML2000}
R.~E. Schapire and Y.~Singer, ``Boostexter: {A} boosting-based system for text
  categorization,'' \emph{Machine Learning}, vol.~39, no. 2/3, pp. 135--168,
  2000.

\bibitem{Snoek:VideoSegment2006}
C.~Snoek, M.~Worring, J.~C. van Gemert, J.~Geusebroek, and A.~W.~M. Smeulders,
  ``The challenge problem for automated detection of 101 semantic concepts in
  multimedia,'' in \emph{ACM MM}, 2006, pp. 421--430.

\bibitem{Snoek:Mediamill2006}
C.~G.~M. Snoek, M.~Worring, J.~C. van Gemert, J.-M. Geusebroek, and A.~W.~M.
  Smeulders, ``The challenge problem for automated detection of 101 semantic
  concepts in multimedia,'' in \emph{ACM MM}, 2006, pp. 421--430.

\bibitem{Tai:PLST2012}
F.~Tai and H.-T. Lin, ``Multilabel classification with principal label space
  transformation,'' \emph{Neural Computation}, vol.~24, no.~9, pp. 2508--2542,
  2012.

\bibitem{Tan:CVPR2015}
M.~Tan, Q.~Shi, A.~van~den Hengel, C.~Shen, J.~Gao, F.~Hu, and Z.~Zhang,
  ``Learning graph structure for multi-label image classification via clique
  generation,'' in \emph{CVPR}, 2015, pp. 4100--4109.

\bibitem{Trohidis:Emotions2008}
K.~Trohidis, G.~Tsoumakas, G.~Kalliris, and I.~Vlahavas, ``Multi-label
  classification of music by emotion,'' \emph{EURASIP Journal on Audio, Speech,
  and Music Processing}, vol.~4, no.~1, pp. 325--330, 2008.

\bibitem{Tsoumakas:BR2010}
G.~Tsoumakas, I.~Katakis, and I.~P. Vlahavas, ``Mining multi-label data,'' in
  \emph{Data Mining and Knowledge Discovery Handbook, 2nd ed.}, 2010, pp.
  667--685.

\bibitem{Tsoumakas:RAKEL2007}
G.~Tsoumakas and I.~P. Vlahavas, ``Random \emph{k} -labelsets: An ensemble
  method for multilabel classification,'' in \emph{ECML}, 2007, pp. 406--417.

\bibitem{Maaten:tSNE2008}
L.~van~der Maaten and G.~E. Hinton, ``Visualizing high-dimensional data using
  t-sne,'' \emph{JMLR}, vol.~9, pp. 2579--2605, 2008.

\bibitem{Vasisht:MIML2014}
D.~Vasisht, A.~C. Damianou, M.~Varma, and A.~Kapoor, ``Active learning for
  sparse bayesian multilabel classification,'' in \emph{KDD}, 2014, pp.
  472--481.

\bibitem{Wang:Sparse2009}
C.~Wang, S.~Yan, L.~Zhang, and H.~J. Zhang, ``Multi-label sparse coding for
  automatic image annotation,'' in \emph{CVPR}, 2009, pp. 1643--1650.

\bibitem{Wang:CNN-RNN2016}
J.~Wang, Y.~Yang, J.~Mao, Z.~Huang, C.~Huang, and W.~Xu, ``Cnn-rnn: A unified
  framework for multi-label image classification,'' in \emph{CVPR}, 2016, pp.
  2285--2294.

\bibitem{Wei:HCP2014}
Y.~Wei, W.~Xia, J.~Huang, B.~Ni, J.~Dong, Y.~Zhao, and S.~Yan, ``{CNN:}
  single-label to multi-label,'' \emph{CoRR}, vol. abs/1406.5726, 2014.

\bibitem{Winn:MSRC2005}
J.~Winn, A.~Criminisi, and T.~Minka, ``Object categorization by learned
  universal visual dictionary,'' in \emph{ICCV 2005}, pp. 1800--1807.

\bibitem{Zhou:Metric2016}
X.~Wu and Z.~Zhou, ``A unified view of multi-label performance measures,''
  \emph{CoRR}, vol. abs/1609.00288, 2016.

\bibitem{Yu:LEML2014}
H.~Yu, P.~Jain, P.~Kar, and I.~S. Dhillon, ``Large-scale multi-label learning
  with missing labels,'' in \emph{ICML}, 2014, pp. 593--601.

\bibitem{Zhang:SIGKDD2010}
M.~Zhang and K.~Zhang, ``Multi-label learning by exploiting label dependency,''
  in \emph{ACM SIGKDD}, 2010, pp. 999--1008.

\bibitem{Zhang:MLKNN2007}
M.~Zhang and Z.~Zhou, ``{ML-KNN:} {A} lazy learning approach to multi-label
  learning,'' \emph{Pattern Recognition}, vol.~40, no.~7, pp. 2038--2048, 2007.

\bibitem{Zhang:Review2014}
M.~L. Zhang and Z.~H. Zhou, ``A review on multi-label learning algorithms,''
  \emph{TKDE}, vol.~26, no.~8, pp. 1819--1837, 2014.

\bibitem{Zhang:ICML2012}
Y.~Zhang and J.~G. Schneider, ``Maximum margin output coding,'' in \emph{ICML},
  2012, pp. 1575--1582.

\bibitem{Zhao:MRV2015}
F.~Zhao and Y.~Guo, ``Semi-supervised multi-label learning with incomplete
  labels,'' in \emph{IJCAI}, 2015, pp. 4062--4068.

\bibitem{Zhou:ML2012}
T.~Zhou, D.~Tao, and X.~Wu, ``Compressed labeling on distilled labelsets for
  multi-label learning,'' \emph{ML}, vol.~88, no. 1-2, pp. 69--126, 2012.

\bibitem{Zhu:Spatial2017}
F.~Zhu, H.~Li, W.~Ouyang, N.~Yu, and X.~Wang, ``Learning spatial regularization
  with image-level supervisions for multi-label image classification,'' in
  \emph{CVPR}, 2017, pp. 5513--5522.

\end{thebibliography}

\end{document}